\documentclass[pdflatex,sn-mathphys-ay, iicol]{sn-jnl}% Math and Physical Sciences 
\usepackage{natbib}
\usepackage{amssymb}
\usepackage{amsmath}
\usepackage{graphicx}
\usepackage{longtable}
\usepackage[table]{xcolor}
\usepackage{pifont}
\usepackage{multirow}
\usepackage{makecell}
\usepackage{booktabs}       % professional-quality tables
\usepackage[utf8]{inputenc} % allow utf-8 input
\usepackage[T1]{fontenc}    % use 8-bit T1 fonts
\usepackage{hyperref}       % hyperlinks
\usepackage{xspace}
\usepackage{url}            % simple URL typesetting

\usepackage{amsfonts}       % blackboard math symbols
\usepackage{amsmath}
\usepackage{nicefrac}       % compact symbols for 1/2, etc.
\usepackage{microtype}      % microtypography
\usepackage{xcolor}         % colors
\usepackage[dvipsnames]{xcolor}
\usepackage{float}
\usepackage{caption}
\usepackage{subcaption}
\usepackage{cleveref}
\usepackage{pifont}
\usepackage{fontawesome5}
\usepackage{makecell}
\usepackage{pifont}

\begin{document}
\newcommand{\soa}{state-of-the-art\xspace}
\newcommand{\diff}[2]{\frac{\partial #1}{\partial #2}}
\newcommand{\vct}[1]{\ensuremath{\boldsymbol{#1}}}
\newcommand{\mat}[1]{\ensuremath{\mathbf{#1}}}
\newcommand{\set}[1]{\ensuremath{\mathcal{#1}}}
\newcommand{\con}[1]{\ensuremath{\mathsf{#1}}}
\newcommand{\tsum}{\ensuremath{\textstyle \sum}}
\newcommand{\T}{\ensuremath{\top}}
\newcommand{\mycomment}[1]{\textcolor{red}{#1}}
\newcommand{\ind}[1]{\ensuremath{\mathbbm 1_{#1}}}
\newcommand{\Ind}{\ensuremath{\mathbb{I}}}
\newcommand{\argmax}{\operatornamewithlimits{\arg\!\max}}
\newcommand{\erf}{\text{erf}}
\newcommand{\sign}{\text{sign}}
\newcommand{\argmin}{\operatornamewithlimits{\arg\!\min}}
\newcommand{\bmat}[1]{\begin{bmatrix}#1\end{bmatrix}}
\newcommand{\myparagraph}[1]{\noindent \textbf{#1}.}
\newcommand{\ie}{i.e.,\xspace}
\newcommand{\eg}{e.g.,\xspace}
\newcommand{\etc}{etc.\xspace}
\newcommand{\aka}{a.k.a.\xspace}
\newcommand{\wrt}{w.r.t.\xspace}
\newcommand{\etal}{et al.\xspace}
\newcommand{\naive}{na\"ive\xspace}
\newcommand{\base}{baseline\xspace}
\newcommand{\IndState}{\State \hspace{\algorithmicindent}}
\newcommand{\edit}[1]{\textcolor{black}{#1}}
\newcommand{\fillertext}{Lorem ipsum dolor sit amet, consectetur adipiscing elit, sed do eiusmod tempor incididunt ut labore et dolore magna aliqua. Ut enim ad minim veniam, quis nostrud exercitation ullamco laboris nisi ut aliquip ex ea commodo consequat. Duis aute irure dolor in reprehenderit in voluptate velit esse cillum dolore eu fugiat nulla pariatur. Excepteur sint occaecat cupidatat non proident, sunt in culpa qui officia deserunt mollit anim id est laborum.}

% Utils
\newcommand{\vmark}{\textcolor{green!70!black}{\ding{51}}} % ✓
\newcommand{\xmark}{\textcolor{red}{\ding{55}}}

\newcommand{\todo}[1]{\textcolor{red}{TODO: #1}}
\newcommand{\note}[1]{\textcolor{olive}{Note: #1}}
\newcommand{\add}[1]{\textcolor{blue}{#1}}
\newcommand{\del}[1]{\textcolor{red}{#1}}
% LOSS
\newcommand{\Loss}{\ensuremath{\mathcal{L}}\xspace}
\newcommand{\AtkLoss}{\ensuremath{\hat{\Loss}}\xspace}
\newcommand{\loss}{\ensuremath{\ell}\xspace}
\newcommand{\celoss}{\ell_{CE}}

% Datasets
\newcommand{\D}{\ensuremath{\mathcal{D}}} % generic dataset
\newcommand{\Dtr}{\ensuremath{\set D}\xspace}
\newcommand{\Dts}{\ensuremath{\set T}\xspace}
\newcommand{\Dval}{\ensuremath{\set V}\xspace}
\newcommand{\Db}{\mathcal{D}^{(b)}}

% DATA NOTATION
\newcommand{\ypred}{\hat{y}\xspace}
\newcommand{\ppred}{\hat{\vct{p}}\xspace}
\newcommand{\x}{\ensuremath{\vct{x}}\xspace}
\newcommand{\Ntask}{\ensuremath{B}\xspace}
\newcommand{\grad}{\ensuremath{\vct g}\xspace}

\newcommand{\norm}[1]{\left \lVert #1 \right \rVert}

% Training
\newcommand{\lr}{\ensuremath{\eta}\xspace}
\newcommand{\atkstep}{\ensuremath{\rho}\xspace}
\newcommand{\Epochs}{\textit{E}\xspace}

% Distributions
\newcommand{\Uniform}{\ensuremath{U(0,1)}}
\newcommand{\Exp}{\ensuremath{\mathbb{E}}}

% Model architecture
\newcommand{\m}{\ensuremath{f}\xspace}
\newcommand{\fext}{\ensuremath{\phi}\xspace}
\newcommand{\clf}{\ensuremath{g}\xspace}

% Metrics
\newcommand{\ACA}{LCA\xspace}
\newcommand{\AIA}{AIA\xspace}
\newcommand{\AF}{AF\xspace}
\newcommand{\AUC}{AUC\xspace}
\newcommand{\AUPR}{AUPR\xspace}

% Experiments Macros
\newcommand{\Cifarten}{CIFAR10\xspace}
\newcommand{\Cifarhun}{CIFAR100\xspace}
\newcommand{\Timgnet}{TinyImageNet200\xspace}
\newcommand{\CifartenFiveT}{\Cifarhun-5T\xspace}
\newcommand{\CifarhunFiveT}{\Cifarhun-5T\xspace}
\newcommand{\CifarhunTenT}{\Cifarhun-10T\xspace}
\newcommand{\CifarhunTwenT}{\Cifarhun-20T\xspace}
\newcommand{\TimageNetFiveT}{T-Imgnet200-5T\xspace}
\newcommand{\TimageNetTenT}{T-Imgnet200-10T}

% Other
\newcommand{\yes}{\ding{51}}%
\newcommand{\no}{\ding{55}}%

\title{Out-of-Distribution Detection for Continual Learning: Design Principles and Benchmarking}

%%=============================================================%%
%% GivenName	-> \fnm{Joergen W.}
%% Particle	-> \spfx{van der} -> surname prefix
%% FamilyName	-> \sur{Ploeg}
%% Suffix	-> \sfx{IV}
%% \author*[1,2]{\fnm{Joergen W.} \spfx{van der} \sur{Ploeg} 
%%  \sfx{IV}}\email{iauthor@gmail.com}
%%=============================================================%%

\author[1,2]{\fnm{Srishti} \sur{Gupta}}\email{srishti.gupta@unica.it}

\author[1]{\fnm{Riccardo} \sur{Balia}}\email{riccardo.balia@unica.it}

\author[1]{\fnm{Daniele} \sur{Angioni}}\email{daniele.angioni@unica.it}

\author[1]{\fnm{Fabio} \sur{Brau}}\email{fabio.brau@unica.it}

\author[1]{\fnm{Maura} \sur{Pintor}}\email{maura.pintor@unica.it}

\author*[1]{\fnm{Ambra} \sur{Demontis}}\email{ambra.demontis@unica.it}

\author[3]{\fnm{Alessandro Sebastian} \sur{Podda}}\email{sebastianpodda@unica.it}

\author[3]{\fnm{Salvatore Mario} \sur{Carta}}\email{salvatorem.carta@unica.it}

\author[4]{\fnm{Fabio} \sur{Roli}}\email{fabio.roli@unige.it}

\author[1]{\fnm{Battista} \sur{Biggio}}\email{battista.biggio@unica.it}

\affil*[1]{\orgdiv{Department of Electrical and Electronic Engineering}, \orgname{University of Cagliari}, \country{Italy}}

\affil[2]{\orgdiv{Department of Computer, Control and Manag. Eng.}, \orgname{Sapienza University},  \city{Rome}, \country{Italy}}

\affil[3]{\orgdiv{Department of Mathematics and Computer Science}, \orgname{University of Cagliari}, \country{Italy}}

\affil[4]{\orgdiv{Department of Informatics, Bioengineering, Robotics and Systems Engineering}, \orgname{University of Genova}, \country{Italy}}

% \affil*[1]{\orgdiv{Department}, \orgname{Organization}, \orgaddress{\street{Street}, \city{City}, \postcode{100190}, \state{State}, \country{Country}}}

%% Abstract
% Please provide an abstract of 150 to 250 words. The abstract should not contain any undefined abbreviations or unspecified references
\abstract{In dynamic real-world environments, machine learning systems must adapt continuously as new data arrives while reliably detecting novel, previously unseen inputs, \eg samples from a new class. 
Continual learning (CL) and out-of-distribution (OOD) detection address these challenges separately, yet their integration is inherently challenging: CL mechanisms designed to preserve past knowledge can significantly alter the representations employed by the OOD detectors. 
While various surveys and benchmarks have examined these domains independently, how these methods should be adapted to work together has not been systematized yet, and it is unclear which combinations yield promising results. These literature shortcomings hinder the progress in addressing these challenges jointly.
In this work, we present a comprehensive and unified taxonomy of existing methods in both the CL and OOD domains, highlighting how out-of-distribution techniques should be adapted to be integrated into continual learning pipelines. Furthermore, we introduce a benchmark to systematically evaluate the performance of state-of-the-art OOD methods within varying continual learning settings. Our results highlight which method combinations are most promising, offering actionable insights for advancing robust, adaptable AI systems, along with guidelines for practitioners.}

%% Keywords
% Please provide 4 to 6 keywords which can be used for indexing purposes.
\keywords{out-of-distribution detection, continual learning, open-world learning, outlier detection}

\maketitle
% %added for arXiv version
\footnotetext{This manuscript is under review at the International Journal of Computer Vision (IJCV).}

\section{Introduction}
Recent years have witnessed significant progress in the development of machine learning models across a wide range of fields, fueled by increased computational resources, large-scale datasets, and the rise of deep learning architectures. From malware detection to enabling autonomous navigation, modern machine learning systems have demonstrated remarkable capabilities. However, as these models are deployed in ever-changing real-world scenarios, their ability to remain reliable and adaptive over time becomes increasingly important. For example, in the real world, new malware families are continuously developed, whereas autonomous driving cars are employed in many different cities and weather conditions. Models trained in fixed settings can not respond effectively to novel conditions encountered post-deployment. In fact, most machine learning models are still developed under the assumption that training and test data are independent and identically distributed (i.i.d.), i.e., sampled from the same underlying (unknown) distribution. While this assumption simplifies model development and evaluation, it does not hold in many real-world applications, where data changes over time and unexpected inputs frequently occur. Retraining models from scratch whenever new data appears is computationally expensive, time-consuming, and impractical in resource-constrained environments. These limitations underscore the need for \textit{Continual Learning (CL)}, which enables models to incrementally learn from evolving data streams without forgetting past knowledge, and \textit{Out-of-Distribution (OOD) detection}, which allows systems to identify and respond to novel or anomalous inputs. Jointly addressing both challenges is critical to developing robust, efficient, and adaptive AI systems. 

\begin{figure*}
    \centering
    \includegraphics[width=0.95\linewidth]{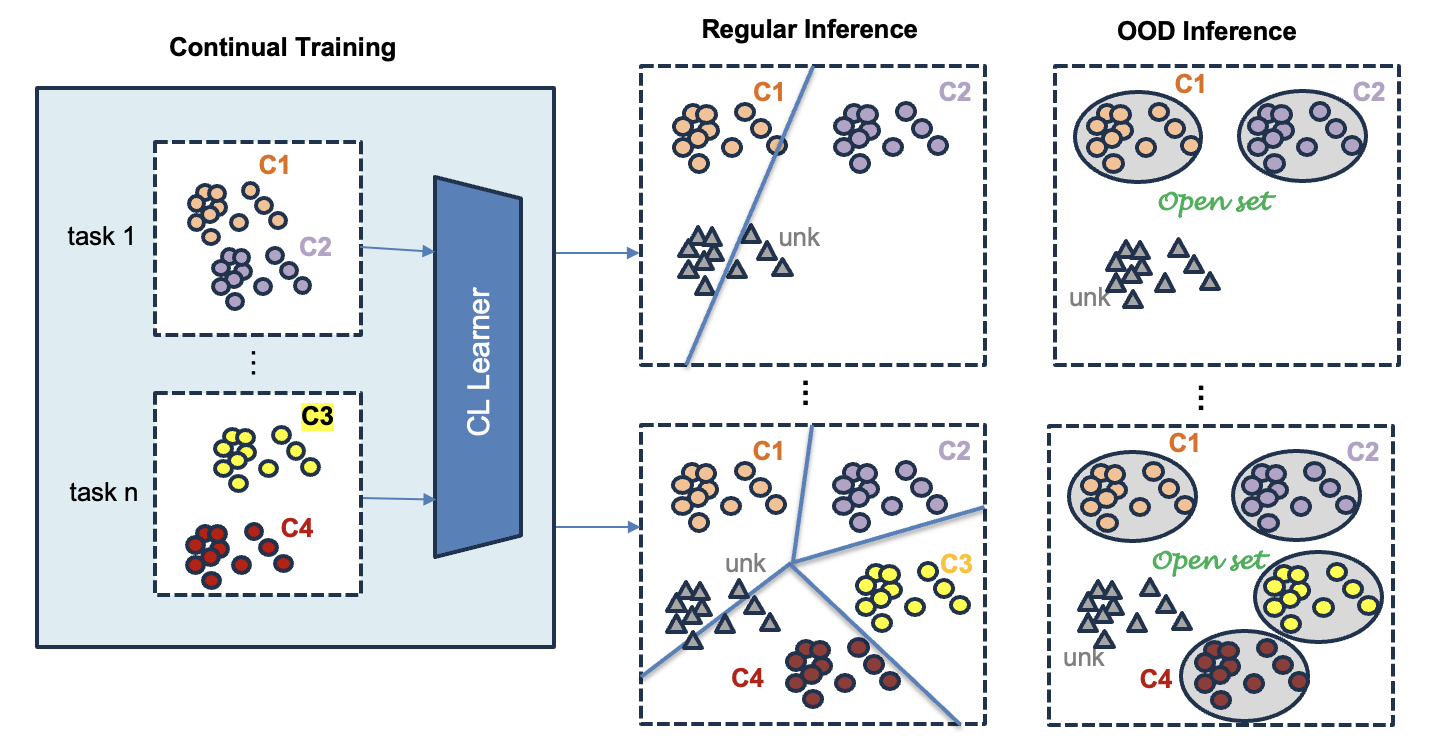}
    \caption{Conceptual Representation of a general post-hoc Out-of-Distribution method employed on a continuously trained model. In the absence of OOD detection, OOD samples: \textit{unk} here, are misclassified into existing classes.}
    \label{fig:enter-label}
\end{figure*}

% Various literature and surveys have divided these domains into different and overlapping categories, making it often difficult to assign a method to a given category.
While existing surveys and benchmarks have addressed CL and OOD independently, there is a lack of understanding regarding how these methods can be effectively combined.
An early attempt in this direction from \cite{miao2026opencil}, benchmarks quite a small set of CL approaches combined with OOD detection methods. While it provides initial evidence that OOD performance degrades as CL accuracy decreases, its scope remains limited, as their analysis focuses on bias toward newly introduced classes and OOD samples, upon which they design an ad-hoc CL–OOD combination, without providing a structured framework to understand the broad interactions across method categories.
The benchmark treats OOD classes in an equal ratio with IND data, which is an inflated scenario in real world setting.

%and do not consider in it into future training stages, limiting its ability to capture real-world CL dynamics.
%
To address this gap, we first categorize both the CL and OOD literature techniques into three meta-categories, for which we provide the formalization, and we present a comprehensive taxonomy in \autoref{sec:background}.
% For CL, the three meta-categories proposed are: Memory-based, Constraint-based and Architecture-based. For OOD, the proposed meta-categories are \textit{adhoc} Training-based, \textit{post-hoc} Calibration-based, and \textit{post-hoc} Inference-based methods.
%These meta-categories are then formalized and incorporate further sub-categories  to cover most methods in the literature. 
% After establishing the taxonomy, we propose a benchmark for adaptive systems that integrate OOD detection with CL.
After establishing the taxonomy, in \autoref{sect:methodology} we derive design principles that clarify how OOD detection methods should be adapted and integrated into CL pipelines, and then propose a benchmark for systematically evaluating such CL–OOD combinations.
%
% Battista: I don't get this part.
%Furthermore, we perform model training using both OOD and CL methods, in other words, OOD-aware CL training, and compare it with integrated methods where the training is done via CL methods and OOD detection is added post-hoc after model training. To our knowledge, we are the first to perform OOD-aware CL training and evaluate it comprehensively. 
%
% \autoref{tab:cl_ood_taxonomy} shows the CL and OOD methods considered in this work. 
% Our empirical findings offer valuable insights for researchers and practitioners seeking to build AI systems that are both adaptable and resilient to distributional shifts. 
Using this benchmark, we report in \autoref{sect:exp} a large-scale empirical evaluation of state-of-the-art CL and OOD methods, systematically analyzing over 450 interactions across different continual learning settings. Our empirical findings reveal consistent patterns and trade-offs in how CL training dynamics affect OOD detection performance, offering actionable insights for researchers and practitioners seeking to build AI systems that are both adaptable and resilient to distributional shifts.

In summary, this work provides the following contributions:
\begin{itemize}
    \item We present a unified taxonomy for CL and OOD that sheds light on the current state of the literature of the respective domains. To this end, we define three main meta-categories for both CL and OOD detection, with further sub-categories that cover the existing literature.
    \item We provide practical guidelines to integrate these CL and OOD detection meta-categories.
    \item We propose a benchmarking protocol to systematically integrate and evaluate a given CL-OOD combination.
    \item We present a large-scale evaluation, where we benchmark more than 450 CL-OOD combinations.
    \item We then provide insights on how OOD detection methods can be adapted for integration within CL pipelines.
%    \item We also perform OOD-aware training in CL scenarios and comprehensively evaluate model training capable of both CL and OOD.% from the get go.
    \item We will open-source the code upon acceptance.
    % \item We will release a public leaderboard.
\end{itemize}

% In our experimental setting, we considered future classes as OOD samples -- a more challenging setting where the OOD samples are near-OOD.

% \mycomment{Budget 1 page - Here we should give a bit of context, explain briefly which is the problem of the state of the art and then explain what we are doing in this work with the reference to the different sections in the sections order.}

\section{Categorizing CL and OOD Methods}
\label{sec:background}
% \mycomment{Budget 9 pages}
% This section categorizes CL and OOD methods according to three high-level meta-categories each (Sects. 2.1-2.2). These meta-categories are functional then to explain how to combine existing OOD and CL methods (Sect. 3). In practice, we set the stage for Section 3.

% PUT THE TABLE WITH THE PAPERS-vs-CATEGORIES HERE and explain that we will walk the reader to the categories in the next sections
In this section, we describe the taxonomy shown in \autoref{tab:supported_methods} and its unified formulation for Continual Learning (\autoref{sec:cl}) and then Out-of-Distribution detection methods (\autoref{sect:ood}). We begin by establishing the notation adopted throughout the paper.

\myparagraph{Notation}
We consider the continual learning setting over a stream of \Ntask tasks $\{\D^{(1)}, \D^{(2)}, \dots, \D^{(B)}\}$, where each task $b$ provides a dataset $\D^{(b)} = \{(\x_i^{(b)}, y_i^{(b)})\}_{i=1}^{n_b}$, with $n_b$ training instances. 
Each input $\x_i^{(b)} \in \mathbb{R}^D$ is associated with a label $y_i^{(b)} \in Y_b$, where $Y_b$ is the set of classes introduced at step $b$.
After training on task $b$, the model is evaluated on the set of all classes seen so far $\set{Y}_b = \bigcup_{j=1}^b Y_j$.
For a given dataset $\mathcal{D}$ we defined the empirical risk loss as:
\begin{equation}
    \mathcal{L}_{\mathcal{D}}(\theta) = \mathbb{E}_{(\vct x,y)\sim \mathcal{D}} \left[\loss(f(\vct x\,;\theta),y)\right],    
\end{equation}
which is a function on the model's parameter, where \loss is a standard classification loss, \eg the cross-entropy loss.
However, in continual learning, the objective is not only to minimize the loss on the current task $b$, but also to keep the risk low on all previously seen tasks, \ie minimizing $\sum_{t=0}^b \mathcal{L}_{\mathcal{D}^{(t)}}(\theta)$.
As we will explore in the next sections, $\mathcal{L}$ can be defined based on the underlying methodology used to overcome CL challenges.

To ease notation, we decompose the classification function into two parts: (i) the feature extractor $h(\cdot\,; \phi)$, parameterized by $\phi$, and (ii) the classification head $g(\vct x\,; w) = w^\top \vct x$. Thus, $f(\vct x\,;\theta) = w^\top h(\vct x\,;\phi)$, where $\theta=(\phi,w)$ represents the complete set of model parameters. We assume the output scores $f(\vct x\,;\theta)$ are unnormalized logits, from which one can obtain the probability vector %$\vct{p} 
$p(\vct x;\theta)= \sigma(f(\vct x;\theta))$, where $\sigma$ is the softmax function. Consequently, the predicted label is denoted as $\ypred = \argmax_{j \in \set{Y}_b} f(\vct x\,;\theta)_j$.

\begin{table*}[t]
% \parbox{.45\textwidth}{
\centering
\rowcolors{2}{white}{gray!10}
\resizebox{\textwidth}{!}{
\begin{tabular}{lccc}
\toprule
\textbf{CL Method} & \texttt{Memory-based (M)} & \texttt{Regularization-based (R)} & \texttt{Architecture-based (A)} \\
% \textbf{CL Method} & \texttt{R} & \texttt{C} & \texttt{A} \\
\midrule
ER~\citep{rolnick2019er_cl} & \yes & & \\ 
MIR~\citep{aljundi2019mir_cl} & \yes & & \\ 
GDumb~\citep{prabhu2020gdumb_cl} & \yes & & \\ 
ER-ACE~\citep{caccia2021erace_cl} & \yes & & \\ 
RAR~\citep{kumari2022rar_cl} & \yes & & \\ 
IL2M~\citep{belouadah2019il2m_cl} & \yes & & \\ 
DER~\citep{buzzega2020dark_cl} & \yes & (\yes) & \\ 
BiC~\citep{wu2019bic_cl} & \yes & (\yes) & \\ 
iCaRL~\citep{rebuffi2017icarl_cl} & \yes & (\yes) & \\
LwF~\citep{li2017lwf_cl} &  & \yes & \\ 
EWC~\citep{Kirkpatrick2017ewc_cl} & & \yes & \\ 
SI~\citep{zenke2017si_cl} & & \yes & \\ 
GEM~\citep{lopez2017gem_cl} & (\yes) & \yes & \\
A-GEM~\citep{chaudhry2019agem_cl} & (\yes) & \yes & \\
Dynamic-ER~\citep{yan2021dynamicER_cl} & (\yes) & (\yes) & \yes \\ 
% \end{tabular}
% }
% \caption{Supported Continual Learning (CL) methods.}
% }
% %
% \parbox{.45\textwidth}{
% \centering
% \rowcolors{2}{gray!10}{white}
% \resizebox{\linewidth}{!}{
% \begin{tabular}{lccc}
\toprule
\toprule
% \textbf{OOD Method} & \texttt{P-TF} & \texttt{P-DT} & \texttt{T} \\ 
\textbf{OOD Detection Method} & \texttt{Training-Time (T)} & \texttt{Calibration-Time (C)} & \texttt{Inference-Time (I)} \\ 
\midrule
OE~\citep{hendrycks2019oe_ood} & \yes &  &  \\
LogitNorm~\citep{wei2022logitnorm_ood} & \yes &  & \\
PixMix~\citep{hendrycks2022pixmix_ood} & \yes &  &  \\

SHE~\citep{zhang2023she_ood} &  & \yes &  \\
Mahalanobis Distance (MD)~\citep{lee2018md_ood} &  & \yes &  \\
ViM~\citep{wang2022vim_ood} &  & \yes &  \\
OpenMax~\citep{bendale2016openmax_ood} &  & \yes &  \\
ReAct~\citep{sun2021react_ood} & & (\yes) & \yes \\
DICE~\citep{sun2022dice_ood} &  & (\yes) & \yes \\
Temperature Scaling (TS)~\citep{guo2017tempscale_ood} &  & (\yes) & \yes \\

MSP~\citep{hendrycks2017msp_ood} & &  & \yes \\
Entropy~\citep{macedo2021entropy_ood} &  &  & \yes \\
EBO~\citep{liu2020energy_ood} &  &  & \yes \\
MaxLogit~\citep{hendrycks2022maxlogit_ood} &  &  & \yes \\
ASH~\citep{djurisic2023ash_ood} &  &  & \yes \\
SCALE~\citep{xu2023scale_ood} &  &  & \yes \\
ODIN~\citep{liang2018odin_ood} &  &  & \yes \\
\bottomrule
\end{tabular}
}
\caption{Supported Continual Learning (CL) and Out-of-Distribution (OOD) detection methods. For each method, we mark its main category with \yes, while we indicate with (\yes) the additional components from other categories.}
\label{tab:supported_methods}
\end{table*}

\subsection{Continual Learning (CL)}
\label{sec:cl}

% CL methods are generally categorized into three or four primary meta-categories based on their strategy for balancing stability (retaining old knowledge) and plasticity (learning new knowledge). 
% The most common categorization involves the following three meta-categories: (see TPAMI 2024 \url{https://github.com/LAMDA-CL/CIL_Survey}, which uses data-centric, model-centric, algorithm-centric -- these could be even more appropriate - see also \url{https://www.sciencedirect.com/science/article/pii/S0004370224001735} - how are we novel against them? (this is for related work))

% Continual Learning is a widely studied topic in recent years offering variety of methods to curb catastrophic forgetting \add{, i.e., the tendency of the models to decrease performance on old tasks while learning new ones}. 
Continual Learning (CL), also known as lifelong learning or incremental learning, is a branch of machine learning that focuses on developing models that can learn continuously from a stream of data over time. The core challenge that CL addresses is the \textit{\textbf{stability-plasticity dilemma}}. \textit{Plasticity} refers to a model's ability to acquire and adapt to new information, while \textit{stability} ensures that previously learned knowledge is preserved. Striking a balance between these two aspects is critical as too much plasticity leads to \textit{catastrophic forgetting}, \ie when new learning overwrites past knowledge; whereas too much stability results in an inability to learn new concepts. CL research is typically organized around three main learning settings: a) Task Incremental Learning (TIL), where training happens in sequential batches of tasks and a task-id is given at train and test time; b) Domain Incremental Learning (DIL) where label space remains fixed but input distribution shifts over time, and c) Class Incremental Learning (CIL), the most challenging setting where new classes are introduced over time without task labels. This work focuses on a CIL setting that can be easily scaled to TIL by simply adding the task-id.

While there exists a vast amount of literature in the field and new methods to curb forgetting are discovered regularly, the diverse landscape of methods often poses a challenge of overlapping categories across literature. To bring coherence to these overlapping perspectives, we group CL into 3 meta-categories: memory-, regularization-, and architecture-based methods. They are formally discussed below.
% which reflects fundamental techniques through which models retain or recover past information. 
% We further provide general formulations of these meta-categories to provide a mathematical foundation that allows practitioners to compare methods systematically.

\subsubsection{Memory-based Methods}
These methods mitigate catastrophic forgetting by storing past knowledge and replaying it during later training and inference stages. This "past knowledge" can consist of explicit stored samples, compressed representations such as templates or prototypes, or metadata and statistics used to calibrate predictions.\medskip

\myparagraph{Exploiting memory during training}
Based on how past knowledge during training is used, replay methods are used to continually train; they can be divided into two sub-categories: \textbf{a) Replay-based methods}, where samples from previous tasks are directly replayed in each training. This past knowledge is typically stored in a buffer, which can be either \textit{fixed} or \textit{growing} if the fixed number of samples per task is stored and the number of tasks increases over time. For example, \textit{Experience Replay} (ER)~\citep{rolnick2019er_cl} is the simplest and a prevalent strategy in CL that stores raw samples in the buffer $\mathcal{E}^{(b)}\subseteq\cup_{t=0}^{b-1}\mathcal{D}^{(t)}$. Therefore, training sequentially with a buffer requires minimizing the loss: 
\begin{equation}
    \mathcal{L}^{(b)}(\theta) = \alpha\mathcal{L}_{\Db}(\theta) + (1-\alpha)\mathcal{L}_{\mathcal{E}^{(b)}}(\theta)  
    \label{eq:exp_replay}
\end{equation}
deduced from a convex combination of the empirical risk of past and present data, where the convex parameter $\alpha\in[0,1]$ is fixed. Another approach, called \textit{generative replay}, fills the buffer with pseudo-examples $\set{E}$ produced using a generative model~\citep{shin2017dgr_cl, Hu2018OvercomingCF, kemker2018fearnetCL, jiang2021ibdrrCL}. They are then trained using \autoref{eq:exp_replay}. 
Differently from ER, \textit{feature replay} methods store more compact feature representations of past samples in the buffer $\mathcal{E}^{(b)}\subseteq\cup_{t}^{b-1} \mathcal{F}^{(t)}$, hence reducing its memory cost, where $\mathcal{F}^{(t)}=\left\{(h(\vct x;\phi_t),y)\,|\,(\vct x,y)\in\mathcal{D}^{(t)}\right\}$ are features from past tasks~\citep{petit2023fetrilCL}. Also in this scenario, a convex combination of risk on present data and data from the buffer is considered, where the past data are only used to train the classification head, minimizing the following loss:
\begin{equation}
    \mathcal{L}^{(b)}(\theta) = \alpha\mathcal{L}_{\mathcal{D}^{(b)}}(\theta) + (1-\alpha)\mathcal{L}_{\mathcal{E}^{(b)}}(w)
\end{equation}
While most methods consider an offline scenario where the batch of data can be processed multiple times, \textit{online-replay} considers a scenario where the data is seen only once by the model and updates the buffer with every training iteration~\citep{lopez2017gem_cl}. \textbf{b) Bias correction methods}, where models trained sequentially tend to get biased towards recent tasks~\citep{buzzega2020corr_cl}, are course corrected to restore balanced performance across tasks. Unlike rehearsal methods, bias correction methods are applied to the model's output to counteract the shift created by task recency. In this case, past knowledge is preserved in the form of stored statistics like feature means, covariances, class prototypes, etc. For example, Bias Correction (BiC)~\citep{Wu19bic_cl} utilizes a held-out memory set of past samples to train a small calibration layer, which adjusts the logits of old versus new classes using learned scaling and bias parameters. However, IL2M~\citep{belouadah2019il2m_cl}, does not store raw samples extensively, but uses stored statistics like class means, classifier norms from previous tasks. These statistics act as a surrogate replay signal, enabling the classifier to be rescaled to reduce forgetting-induced bias. \medskip

\myparagraph{Exploiting Memory during Inference} After training, \textbf{c) Template-based methods} store compressed representations of past data in the form of class prototypes in the memory. These prototypes are usually defined as the per-class centroids in the (last) feature space and can be denoted as:
\begin{equation}
P_{c} = \frac{1}{N_c}\sum_{(\vct x,y)\in\mathcal{D}^{(b)}}\mathbb{I}(y=c)h(\vct x;\theta),
\end{equation}
where $N_c$ is the number of samples in class $c$. These prototypes are then used at test time for query-template matching, meaning, matching the incoming sample with the nearest prototype, i.e., $y^* = \argmin_{y\in|\set Y_b|}\|h(x;\theta) - P_y\|$. For example, \textit{iCaRL} performs inferences using class prototypes. Some surveys treat template-based methods as a separate category; however, considering the principles of storing past information, we consider it as a memory-based method. Moreover, since continual training causes semantic drift~\cite{yu2020semanticdrift_cl} in embedding space, template-based methods tend to use set of exemplars to re-calculate the prototypes after each stage~\citep{Lange2021prototype_cl}. This is done to ensure compatibility of the prototypes with the newest embedding space and, consequently, mitigate the drift.

% After training, we collect a memory of past information written in terms of $\mathcal{M}^{(b)} = \aleph(\mathcal{F}^{(1)}, \ldots, \mathcal{F}^{(b)})$
% $\{w_1,...,w_C\}$ centroid in the feature space.
% \[
%     g_0(h) = \max_i -\|h-w_i\| = - \max_i \|h-w_i\|
% \]
% \[
% \|w_i-h\|^2 = \|w\|^2 + \|h\|^2 -2w_i^th
% \]
% If $\|w_1\|=1$ and $\|h\|=1$
% \[
% \|w_i-h\|^2 = 2-2w_i^th
% \]
% \[
%     g_1(h) = 2-2*
%     \begin{bmatrix}
%         w_1^t/\|w_1\| \\
%         \vdots\\
%         w_C^t/\|w_C\|
%     \end{bmatrix} \frac{h}{\|h\|}
% \]

% Finally \textbf{c) Template-based methods} where compressed representation templates of past data are stored in the form of prototypes instead of raw samples. They are different from the previous sub-categories in a way that they are instead used at test time for query-template matching, meaning, matching the incoming sample with the nearest prototype. For example, \textit{iCaRL} perform inferences using class prototypes. Some surveys treat bias-correction and template-based methods as a separate category; however, keeping the principles of storing past information, we consider it as a memory-based method.

Memory-based methods are effective for reducing catastrophic forgetting, as these methods have an effective way of preserving past information. The downside to these methods is a requirement for memory storage, which may be a limiting factor when the storage is constrained.

\subsubsection{Regularization-based Methods}
Constraint-based methods prevent forgetting by restricting the model updates via weight penalties, distillation targets, or gradient constraints, preserving important representations without needing to replay past samples. Depending on the model's space being constrained, these methods can be formulated as:
\begin{equation}
    \mathcal{L}^{(b)} (\theta) = \alpha \mathcal{L}_{\Db}(\theta) + (1-\alpha)\mathcal{R}^{(b)}(\theta),
    \label{eq: constraint_loss}
\end{equation}
where the additional loss term $\mathcal{R}$ constraints the model from deviating too far from its previous knowledge.

Most common constraint-based methods are: \textbf{a) Knowledge Distillation} (KD) that regularizes the model to enforce learned representation to be similar to that of the previous model. KD operates in the model's \textit{final output} space, \ie predictions or logits, referred to as ``soft-logits". It enables knowledge transfer via student-teacher model~\citep{hinton2015kdCL}, meaning, enabling old (teacher) model to assist new (student) model in learning tasks and can be represented as: $\mathcal{R}^{(b)}(\theta)=\mathbb{E}_{(\vct x,y)\sim\mathcal{D}^{(b)}} \left[KL(p(\vct x;\theta)|| p_t(\vct x;\theta_{b-1}))\right]$ where the new model is encouraged to have similar output class probability scores $p(\cdot\,;\theta)$ of the previous model $p(\cdot\,;\theta_{b-1})$ by computing the KL divergence between the two distributions. Common methods include LwF~\citep{li2017lwf_cl}, iCaRL~\citep{rebuffi2017icarl_cl}, DMC~\citep{Zhang2020dmc_cl}, etc.
% $\mathcal{R}^{(b)}=\sum_{x\in\cup_{t=1}^{b}D^{t}} KL(p_{b}(\vct x)|| p_b(\vct x))$
% % $\mathcal{L} = \beta \loss_{kd}(z(\x), z^{b-1}(\x))$ 
% where $p$ represents soft-targets.
%
\textbf{b) Parameter Regularization} (PR) operates in the space of the model's \textit{weights}. Here, each parameter's importance to the network is evaluated, and the important ones are constrained to be static to maintain former knowledge, and the less important ones are penalized. These methods can be unifyingly denoted as: $\mathcal{R}^{(b)}= \Sigma_k \Omega_k(\theta^{(k)}_{b-1} - \theta^{(k)}_b)^2$ where $\theta^{(k)}_b$ is the $k$-th model parameters when training task $b$, $\theta^{(k)}_{b-1}$ $k$-th parameter after learning last task, and $\Omega_k$ weighs the parameter shift to ensure important parameters do not shift from the last stage. Some methods in this category are EWC~\citep{Kirkpatrick2017ewc_cl}, SI~\citep{zenke2017si_cl}, MAS~\citep{aljundi2018mas_cl}, etc.
Finally, \textbf{c) Data Regularization}: while these methods overlap with replay-based methods as they need a memory buffer, they fundamentally differ in their use. In fact, the replay data is not learned directly; rather, it's used to put a constraint on the gradients before each optimization step. This ensures that new updates do not completely overwrite parameters important for past tasks. For example, in GEM~\citep{lopez2017gem_cl}, gradients of the loss are calculated w.r.t the past samples stored in data buffer that controls the optimization direction and can be denoted as: $\theta_b =\theta_{b-1} -\beta\Pi_\mathcal{E}\left(\nabla\mathcal{L}^{b}(\theta_{b-1})\right)$ where $\Pi_{\mathcal{E}}$ projects the current gradient based on the buffer's samples gradient. Other methods include A-GEM~\citep{chaudhry2019agem_cl}, Adam-NSCL~\citep{Wang2021adam-nscl_cl}, etc.

Constraint-based methods are good for privacy as they require little or no help from large buffers, but may not prevent forgetting entirely. Strong constraints can sometimes hinder the learning of new tasks.

% Based on how constraints are applied, they can be divided into: \textbf{a. Regularization-Based Methods} where these methods impose constraints or penalties on the model's loss function during training to prevent significant changes to parameters that are crucial for previously learned tasks. They avoid the need to store raw data from past tasks. Some known methods are 

% Key Idea: Identify "important" weights for old knowledge and penalize their modification when learning new information.
% Examples: EWC (Elastic Weight Consolidation), SI (Synaptic Intelligence), LwF (Learning without Forgetting, which uses knowledge distillation as a form of output regularization). 

\subsubsection{Architecture-based Methods}
Architecture-based methods mitigate forgetting by allocating separate or expanded model components for new tasks, preventing interference with previously learned representations. This increases the model capacity over time, though often at the cost of increasingly larger memory requirements.
For a new task $b$, %the architecture change can be seen as $\theta_b = \theta_{shared} \cup \theta_b$, resulting in the following learning objectives, which properly weight the two model components updates:
the loss function involves a penalty term to constrain the parameters of the new model that can be updated. Formally, 
\begin{equation}
    \mathcal{L}^{(b)}  (\theta)= \alpha \mathcal{L}_{\mathcal{D}^{(b)}}(\theta) + (1-\alpha) \mathcal{C}(\theta,\theta_{b-1}),
    \label{eq: arch_cl}
\end{equation}
%loss{i}{theta}
%\data{i}
% where $\mathcal{P}$ is a penalty term over a given constraint $\mathcal{C}(\theta,\theta_{b-1})$ ensuring %efficient resource parameter sharing, i.e., 
where the second term is a penalty term ensuring that only a subset of the parameters of the model can be actually modified (\eg preserving the backbone $\phi_b\equiv\phi_{b-1}$, or a predefined set of parameters $M\odot \phi_{b}\equiv M \odot \phi_{b-1}$ where $M$ is a binary mask.)
Some architecture-based methods protect critical pathways for each task via gating or masking. For example, in PathNet~\citep{fernando2017pathnet_cl}, where parts of the network are selected and frozen for future re-use, in PiggyBack~\citep{mallya2018piggyback_cl}, binary masks are learned across a shared base network to specialize weights per task. Other methods consider expanding the network either by adding new neurons, layers or modules to increase the capacity of the model as new tasks arrive. For example, in PNN~\citep{rusu2022pnn_cl}, where new networks are added per task, and lateral connections are maintained with the previous network for knowledge transfer, whereas in DynamicER~\citep{yan2021dynamicER_cl}, the model grows in both features and classifiers to adapt new tasks without altering previous representations.
% \textbf{a) Parameter Isolation}, where each task uses a dedicated subset of the network parameters, effectively isolating the learning process to prevent interference with previous tasks. Shared layers may exist, but critical pathways for each task is protected via gating or masking mechanisms. For example, in PathNet~\citep{fernando2017pathnet_cl}, where parts of the network are selected and frozen for future re-use, in PiggyBack~\citep{mallya2018piggyback_cl}, binary masks are learned across a shared base network to specialize weights per task.
%
% \textbf{b) Expansion Methods} where when a new task arrives, the architecture expands by adding new neurons, layers, or modules, enabling the model to increase capacity over time. This allows the model to learn new concepts without overwriting previous ones. For example, in PNN~\citep{rusu2022pnn_cl}, where new networks are added per task and lateral connections are maintained with previous network for knowledge transfer, whereas in DER~\citep{yan2021dynamicER_cl}, model grows in both features and classifiers to adapt new tasks without altering previous representations.

These structure-based methods are proven to be very effective in mitigating catastrophic forgetting, even when scaling task-wise. However, some of these methods need task-id at inference time which is an unrealistic assumption. Moreover, irrespective of the presence or absence of the task-id, size of the network can grow quickly as the number of task trainings increase.

% These methods address forgetting by dynamically modifying or expanding the network architecture as new tasks arrive, often by allocating specific sets of parameters or modules to each task. This ensures that learning new tasks does not interfere with the parameters used for old tasks. 
% Key Idea: Isolate parameters to prevent interference between tasks.
% Examples: PNN (Progressive Neural Networks), HAT (Hard Attention to the Task), methods that add new branches or use task-specific prompts, or Dynamic ER, which saves a frozen backbone for each task and further aggregates the features for classification.

\subsection{Out-of-Distribution (OOD) Detection}
\label{sect:ood}

Out-of-Distribution refers to the task of identifying test samples whose underlying data distribution differs from that of the training set. In principle, it aims to quantify whether an input lies within the learned space of the In-Distribution (IND) manifold. The key challenge motivating OOD detection is the overconfidence of neural networks, \ie trained neural networks often produce high confidence predictions for semantically unrelated inputs.
To categorize whether a sample is in or out of distribution, a common strategy adopted in the literature is to define a score $s(\x)$ to detect if the sample lies outside the training distribution, \eg samples from future classes, or drifted samples from the known classes. Based on where in the ML pipeline OOD awareness is incorporated, we categorized the existing methods into three meta-categories: a) Training-based, b) Post-hoc: Calibration Time, c) Post-hoc: Inference Time.

\subsubsection{Inference-Time Post-Hoc Methods}
% These are pure "plug-and-play" methods that operate entirely during testing/inference time. They use the raw outputs of any standard pre-trained model and require no further training, calibration, or access to any ID data beyond the initial training phase.
% Keywords: Pure Post-Hoc, No Calibration, Inference-Only (e.g., MSP, Energy, DICE, ReAct).

These methods treat the trained underlying classifier as fixed and operate post-training to derive an OOD score by considering the model's outputs, thereby separating in-distribution from out-of-distribution inputs. Since these methods are not dependent on the classifier itself, they're often termed as training-agnostic methods in the literature. 
Given a trained model $f(\cdot;\theta)$, 
%let's consider that $f$ is decomposed into  $L$ layers whose outcome is $h(\vct x;\phi)$ so that $f=g\circ h^{(1)}\circ\cdots\circ h^{(L)}$, 
an OOD scoring function $s(\vct x)\in\mathbb{R}$ can be defined by involving outputs or inner activations using a pre defined post-hoc transformation $\Psi$ deducing $s(\vct x)= \Psi\left(\vct x; \theta\right)$, where $\Psi$ is a transformation function that maps model outputs to scores, \eg clipping, scaling, or an identity function that operates directly on activations.
Using this score, we define a sample $x$ out-of-distribution, if the score $s(\vct x)$ exceeds a certain threshold $\tau$.
\begin{equation}
    \text{OOD}=\begin{cases}
    1, & \text{if $s(\vct x) \geq \tau$} \\
    0, & \text{otherwise}
    \end{cases}
    \label{eq: ood_score}
\end{equation}
These methods can be further seen in two groups: 
\textbf{a) Scoring Functions}, where the vanilla model outputs are inputted to the transformation function to get the OOD score. For example, in Maximum Softmax Probability (MSP)~\citep{hendrycks2017msp_ood}, the score is deduced by taking the max on softmax, $\Psi(\vct x;\theta)=\max_i p(\vct x;\theta)_i$; in Max Logit (ML)~\citep{hendrycks2022maxlogit_ood} the maximum of unnormalized logits is considered; while, in Energy Scores~\citep{liu2020energy_ood},  the score is deduced considering  a soft-max on the logits, $\Psi(x;\theta)= -\log \sum \exp(Tf(\vct x;\theta))$, which depends on a temperature $T\ge1$ and that converges to Max Logit when $T\to+\infty$).\medskip

\textbf{b) Output Adjustments}: where the outputs of the models are manipulated to enhance IND-OOD separation. For example, in \textit{ReAct}~\citep{sun2021react_ood} where the features $h(\vct x;\phi)$ are clamped to not exceed a threshold $\tau$, computed on the measured percentiles, deducing $\vct{h}'= \min\{h(\vct x; \phi), \tau\}$. The clamped activations are then used to deduce the logits $\vct z = w^\top \vct h'$ from which the score $s(\vct x)$ can be computed using the aforementioned strategies and \autoref{eq: ood_score} to perform the detection. In \textit{DICE}~\citep{sun2022dice_ood}, sparsification is performed on the weights of the last fully connected layer (i.e., the classification head) based on the measure of their contribution, leveraging a mask $M$ so that logits are deduced by $\vct z = (\mathrm{M} \odot \vct{w})^\top h(\vct x;\phi)$.
% $\mathrm{M}$ such that the new outputs are: 
%$f(\x)=(\mathrm{M} \odot \vct{w})^\top \vct{h} + \vct{b}$. 
Outputs from this function are then used to compute the OOD scores. Note that Output Adjustment methods can be further enhanced using Scoring functions when calculating the OOD scores.

These methods do not require retraining or architectural changes and can be easily integrated, almost in a plug-and-play fashion, on an underlying CL model, making them versatile for already deployed models. However, for these methods to perform well, the prerequisite is for the CL method to be optimal, as these OOD methods can not recover training errors. Since forgetting is an ongoing challenge in CL, the performance of detectors is directly impacted by the CL performance.

\subsubsection{Calibration-Time Post-Hoc Methods}
% These methods are applied after the main training is complete but require a separate, often brief, calibration step using a held-out in-distribution (ID) validation dataset to tune parameters or fit statistical models.
% Keywords: Post-Hoc, Calibration Required, Uses ID Validation Data (e.g., for Mahalanobis fitting, SHE, OpenMax).

These methods capitalize on geometry in the learned feature space. The intuition is that after training, features extracted from IND data tend to cluster around task-specific prototypes or centroids, while OOD samples lie farther away in this embedding space. Unlike generic scoring rules, they use explicit distance or similarity judgments, and therefore, the OOD score $s(\vct x)$ can be calculated using:

\begin{equation}
    s(\vct x) = -\min_{b=1,\ldots, B} \mathbb{D}(h(\vct x), \mu_b)
\end{equation}
where $\mathbb{D}$ is the distance-metric, $\mu_b$ is a representative embedding of $b$-th task. The negative form of the equation is considered, so if the score $s(\x)$ falls below the threshold, the sample is OOD. While these methods operate in the inference stage, the representative feature statistics, used to estimate the detector's parameters, are computed in the post-training stage.

Commonly used distance-based method is Mahalanobis Distance (MD)\citep{lee2018md_ood} where the distance is computed between the test sample $x$ and distribution of task $b$ using: $\mathbb{D}_{md}(\vct x, b)= (f(\vct x;\theta)-\mu_b)^\top \Sigma^{-1}(f(\vct x;\theta)- \mu_b)$. While MD poses parametric constraints, in k-Nearest Neighbor Distance (kNN)~\citep{sun2022knn_ood} a non-parametric approach using local feature density is leveraged for OOD detection where score for test sample $x$ is computed as: $s(\vct x)=-\frac{1}{k}\sum_{i=1}^k||h(\vct x;\phi)- h(\vct x^{*}_i;\phi)||_2$ where $\vct x^{*}_i$ are the $k$-nearest neighbors to $\vct x$ in the last training dataset $\mathcal{D}^{(B)}$. In Virtual-logit Matching (ViM)~\citep{wang2022vim_ood}, a virtual logit is constructed based on the distance of a feature combining information from both the feature space and the logit space of a neural network, and thus the score is calculated as: $s(\vct x)= \max_i(f(\vct x;\theta) + v(\vct x))_i$ where $v(\vct x)$ is the virtual logit.

These methods provide interpretable geometrical reasoning, do not rely on the network architecture or outlier exposure; however, they are sensitive to embedding quality and require storage of feature statistic information. Moreover, with continual updates of the model, the feature statistics tend to drift, which is hard to capture.

\subsubsection{Training-Time Ad-Hoc Methods}
% These methods require modifying the core loss function or model architecture during the initial training phase to explicitly optimize for OOD detection capabilities. They are tightly integrated with the model's learning process.
% Keywords: Ad-Hoc, Training-Driven, Outlier Exposure, Architecture Modification.

Training-based OOD methods modify the learning process itself so that the resulting model becomes intrinsically sensitive to out-of-distribution inputs. Formally, these methods optimize an augmented training objective that reduces overconfidence in unsupported regions of the input space and sharpens the classification manifold for in-distribution data.
The learning objective can be defined as:
%\begin{equation}
%    \tilde{\mathcal{L}}(\theta) = \mathcal{L}_{(\vct x,y)\sim\mathcal{D}_{in}}(f(\vct x;\theta), y) + \mathcal{L}_{aux}(\theta)
%\end{equation}
\begin{equation}
    {\mathcal{L}^{(b)}}(\theta) = \mathcal{L}_{\mathcal{D}_{in}^{(b)}}(\theta) + \mathcal{L}_{sep}(\theta)
\end{equation}
where the first loss function is applied to the dataset $\mathcal{D}^{(b)}_{in}$ of samples categorized as \textit{in-distribution}, and the second loss term $\mathcal{L}_{sep}$ enforces training to internally enforce separation between in-distribution samples from unfamiliar inputs, either by explicitly training on external dataset or implicitly regularizing the model training.
% i.e., defined through an external dataset (supervised) or based on previous detection (self-supervised).

Based on whether the auxiliary OOD data is used at the time of training, these methods can be divided into: 
\textbf{a) With OOD Access}, where the model is exposed to the OOD dataset $D_{out}$ during training alongside IND data to supervise the model’s decision boundary to push OOD regions away from IND manifolds, reducing overconfidence on anomalies and increasing separability. For example, in Outlier Exposure (OE)~\citep{hendrycks2019oe_ood}, the classifier is trained to produce low confidence on auxiliary outliers by encouraging model outputs on $D_{out}$ to match the target ``non-IND" distribution, commonly the uniform distribution over classes. This makes the model less confident in regions not covered by $D_{in}$. In Energy-bounded learning~\citep{liu2020energy_ood}, where the model is fine-tuned to explicitly create an energy gap by assigning lower energies to the in-distribution data, and higher energies to the OOD data. 
\textbf{b) Without OOD Access} methods, where the model is trained only on IND data and yet encouraged to behave cautiously in unsupported regions of the input space. Here, the constraints are imposed during training so that IND predictions remain confident and OOD inputs produce smaller, better-separated scores without the need for extra OOD data. For example, in LogitNorm~\citep{ding2025enhancing_ood}, the logits are normalized during training to mitigate overconfidence in OOD samples. In Augmix~\citep{hendrycks2020augmix_ood}, stochastic mixed augmentations are created by composing simple augmentations (\eg rotate, translate, etc.) into chains and mixing them with augmented variants to form training inputs.

% \begin{figure*}[ht]
%     \centering
%     \includegraphics[width=0.95\linewidth]{clood_library.pdf}
%     \caption{CLOOD library structure.}
%     \label{fig:benchmark_structure}
% \end{figure*}

\section{Design Principles and Benchmarking} 
\label{sect:methodology}
% In this section, we discuss the basic design principles for combining the three meta-categories of CL with the three meta-categories of OOD. We then discuss our benchmarking paradigm and the evaluation protocol considered to assess the performance of the combination methods.
In this section, we first outline in \autoref{sect:design} the theoretical design principles underlying CL–OOD integration, and then describe our benchmarking protocol in \autoref{sect:bench}.

\subsection{Design Principles}
\label{sect:design}
% \mycomment{Budget 2 pages - The goal of this section is to explain the basic design principles for combining the 3 categories of CL with the 3 cathegories of OOD. From this section, it should be clear also what we expect to see from the experimental results, in terms of which hypotheses... e.g. we should expect this OOD category to work best when paired with this CL category, etc. see the listing below.}
% \textcolor{blue}{
% Bat: First we need to reason on the 3x3 combinations. I make few points below. We have data/model/algo vs training/calibration/inference. I'd like to put OOD as the primary categories here... see the examples in the list below. A nice accompanying 3x3 table should also clarify the points.}
%
We consider OOD meta-categories as the primary axis and analyze their interaction with CL categories across memory-, regularization- and architecture-based methods.

\myparagraph{Inference-time OOD} These methods are largely orthogonal to the CL mechanisms and can be applied with minimal interference. With memory- and regularization-based methods, these OOD detection methods are fully compatible and operate directly on the model's output. For architecture-based detection methods, they can be applied per head or using the shared backbone.
% no problem at all. This setting is largely orthogonal to the CL mechanism and can be applied with minimal interference, as it does not require modifications to either the training dynamics or the model parameters.

\myparagraph{Calibration-time OOD} This setting requires fitting an OOD detector post each training. They rely on representative features of individual tasks after their training. These detectors strongly rely on a fixed feature representation, which is the case for a non-CL scenario, but highly challenging in CL, where training on a new task inherently comes with a distribution drift in the feature space. For this reason, these detectors must be calibrated after learning each task to function correctly.
% Intuitively, in a continual training setup, the model is updated incrementally and detectors are sensitive to activation shift, which is not the case in standard static setting. 
In both memory-based and regularization-based methods, the feature representation can change substantially, reason for which we expect calibration-based detectors to be more challenging to integrate.
% detection is challenging. Here, shared representations keep changing across tasks, creating activation drift that degrades calibration. 
This is a direct trade-off of preserving past information versus meaningful homogeneous centroids of tasks to compute distance. However, in architecture-based methods, the isolated parameter per task stabilizes feature space, making it a potential match for post-hoc calibration-time detectors.
% 
% This setting requires fitting a post-hoc OOD detector. This approach is expected to be effective in data-centric CL methods, as the availability of replay data or memory buffers enables incremental re-training and stable calibration of the OOD detector (provided that the OOD detector can be trained INCREMENTALLY!). It is also well-suited to model-centric CL approaches that enforce parameter isolation (e.g., frozen classifier heads), since the stability of the representation space supports consistent post-hoc calibration (even if not incremental). In contrast, its effectiveness may be limited in algorithm-centric CL methods, where continual updates to the shared representation induce significant activation drift, making it difficult to maintain a well-calibrated OOD detector over time.
% \textcolor{blue}{DA: These methods can work in two ways: (i) \textit{with replay data}, where, if available, can be used to finetune past representations, or/(ii) \textit{without replay data}, where the feature representation is required to be fixed, or with very minimal changes across different tasks. Using the replay data to update the detectors incrementally can be effective, but it requires defining proper update rules, which are currently done with heuristics. If replay data is not available, instead, the CL method must guarantee that the statistics needed for the detector remain compatible across different tasks.}

\myparagraph{Training-time OOD} These methods incorporate OOD awareness directly during training. 
When integrated in CL, the resulting optimization process requires jointly addressing incremental learning, forgetting prevention, and in-distribution–OOD separation.
% To perform this with CL, the training process is now adjusted to suit three objectives: 
% train incrementally, prevent forgetting, and add OOD awareness. While the first objective is standard training, the second objective corresponds to the CL method used, and the third objective helps maintain IID-OOD separation. 
Training using such methods can be done either \textit{with OOD} data or \textit{without OOD} data. Former case of using additional OOD data would benefit from integration with memory-based methods, as this will provide wholistic understanding data: past data from replay buffers and unknown data from additional OOD data. On the flip side, it also means that they tend to be resource and compute heavy. Additional data may also benefit regularize the model to achieve all three objectives but over constraining can also compete with stability-plasticity dynamics. Finally, integrating additional OOD data learning with architecture methods requires training individual heads or a subset of the model parameters. Structural stability of isolated parameters could potentially reduce interference from . In the second case of no additional OOD data, this setting typically introduces an additional regularization objective during CL. As a result, it can be integrated relatively seamlessly with most CL paradigms. 
    
% Since the model training is regularized for continual updates of new information and particularly memory-based approaches (\eg replay or buffer-based methods), where OOD-aware regularizer can be naturally incorporated into the existing training loop. (is this true? any other consideration vs model-centric and algorithm-centric?)
% \textcolor{blue}{DA: from pipeline perspective (ignoring all bugs encountered when trying to do so) they are indeed easy to integrate, especially if they only require a slight adjustment like LogitNorm, or additional data such as PixMix. In terms of efficacy, instead, this is another story: in OE the new regularizer might interfere too much with the already complex CL mechanism (especially in constraint-based methods, which already perform poorly). PixMix instead, which only improves generalization, is directly compatible and always lead to improvements over the baseline.}

\subsection{Benchmarking}
\label{sect:bench}
% \mycomment{Budget 2 pages - The goal of this section is to explain the methodology that we have employed in our benchmark}
We now describe the end-to-end pipeline of our benchmarking procedure, given a (i) CL method $C$, (ii) a training-time OOD detector $D$, and a (iii) post-hoc OOD detector $d$.
The framework integrates Avalanche as the CL infrastructure \citep{lomonaco2021avalanche} and PyTorch-OOD for OOD detection \citep{kirchheim2022pytorchood}, using the Avalanche plugin-based design that enables OOD modules to act at well-defined steps, \eg after forward pass, or before backpropagation.
% The benchmark proposed in this work provides a unified framework for evaluating continual learning methods that learn new tasks incrementally while also integrating OOD-aware strategies and detectors to recognize never-before-seen ones (\eg new classes). 
Our procedure follows three main stages, which are repeated for each task: (i) \textit{training}, (ii) \textit{calibration}, and (ii) \textit{evaluation}, each defining how $C$, $D$ and $d$ interact within the benchmark.
\medskip

\myparagraph{Training}
Before adapting $D$, we first ensure each of its step-by-step components does not interfere with the procedure defined by $C$. If there is no interference, the two methods can be seamlessly integrated using the plugin-based design. Otherwise, we introduce a lightweight adaptation, either by reordering operations (\ie for a given step, we decide who goes first between the CL or training-tim OOD module) or by modifying $D$ to eliminate the interference.
% We perform this analysis to ensure each training-time OOD method has a single adaptation that makes it compatible with any CL method.
This analysis led us to the following criteria, based on whether $D$ requires access to outlier data or not.
Methods \textit{with OOD access} typically combine the standard classification loss with an additional outlier-based loss term, \eg $\Loss(\theta) = \Loss_{IND}(\theta) + \lambda \Loss_{OOD}(\theta)$. To integrate it with $C$, we simply set $\Loss_{IND}(\theta)$ to the task-specific CL loss $\Loss^{(b)}(\theta)$, ensuring that the original CL objective remains intact.
Methods \textit{without OOD access} operate exclusively on in-distribution data and are therefore easily integrated as follows.
Augmentation-based approaches, such as AugMix or PixMix, are directly applied to both current samples and replayed samples.
Instead, representation-level adjustments, such as LogitNorm, are inserted, without specific adaptations, before the corresponding training step so that $C$ optimizes over the transformed representations.
We note that each method $D$ in our benchmark is adapted to be compatible with any CL strategy. If a new CL method $C$ causes specific interference with the approach $D$ currently in use, it is adapted in an ad-hoc manner.
\medskip

\myparagraph{Calibrating}
This stage calibrates post-hoc, data-dependent detectors so that they operate optimally on each task.
After training a given pair $(C, D)$, we select a post-hoc OOD detector $d$, and we calibrate it using data from the current task $\D^{(b)}$.  
We adapt detectors that require per-task hyperparameter tuning, such as ReAct or Temperature Scaling, by simply overwriting these parameters during each calibration step.
Instead, detectors that rely on estimating task or class-level statistics (\eg Mahalanobis distance or SHE) require a different adaptation. Since these methods were originally designed to compute statistics using the full dataset at once, we modify them to update the statistics of the newly encountered task without recomputing those from previous tasks, following the incremental procedure suggested by \cite{lee2018md_ood}.
Although the underlying principle remains unchanged (most methods rely on class prototypes or class-conditional statistics), some detectors require specific adjustments to support incremental updates, for example, Mahalanobis must update both the class means and the covariance matrix as new data become available.
\medskip

\myparagraph{Evaluating}
To assess the overall performance, we evaluate both the continual learning behavior of the chosen method (with or without a training-time OOD component) and the OOD detection effectiveness of the post-hoc detector applied to the final model.
Let's first define as $\set{A}_j^i$ the accuracy on task $j$ after training task $i$. This allows us to compute the following three standard metrics:
\begin{itemize}
\item \textit{Average Classification Accuracy} (ACA) at task $t$, \ie $\text{ACA}_t = \frac{1}{t}\sum_{j=1}^t \set{A}_j^t$, where for the sake of readibility we refer to $\text{ACA}$ as average classification accuracy after the last task $T$ (\ie  $\text{ACA} = \text{ACA}_T$).
\item \textit{Average Incremental Accuracy} (AIA), \ie $\text{AIA}=\frac{1}{T}\sum_{t=1}^T \text{ACA}_t$, which further describes the overall incremental process by averaging how the accuracy evolved.
\item \textit{Average Forgetting} (AF), which measures how much performance on past tasks deteriorates as new ones are learned, as detailed by \cite{chaudhry2018rwalk_cl}.
\end{itemize}
To evaluate OOD detection performance, tasks up to $t$ are treated as in-distribution (IND), whereas tasks beyond $t$ are considered OOD (\ie never-before-seen classes). Given a detector that assigns a score to each sample in the whole dataset, we compute standard detection metrics, namely AUROC and FPR\@95TPR. The final OOD performance is obtained by averaging these metrics across all tasks.

\subsubsection{Benchmark Setup}
Having described the benchmarking pipeline, we now detail the specific CL and OOD methods, datasets, and training configurations used in our experiments.

\myparagraph{Supported CL and OOD Detection Methods}
The continual learning and OOD-detection algorithms supported by our benchmark are summarized in \autoref{tab:supported_methods}. Overall, we benchmark 15 CL methods and combine them with 14 OOD detectors (8 data-agnostic and 6 data-dependent), resulting in 210 CL–OOD configurations.
In addition, we employ 3 training-time OOD methods coupled with 6 among the best performing CL methods per category (\ie BiC, iCaRL, ER-ACE, Replay, LwF and DynamicER), yielding 18 additional incremental models, each further combined with the 14 OOD detectors (\ie 252 additional CL–OOD configurations).
The benchmark is designed to be easily extensible, allowing new methods to be incorporated seamlessly and evaluated under the same protocol.
\medskip

\myparagraph{Datasets}
We conduct our experiments on the CIFAR-100 dataset \citep{krizhevsky2009learning}, containing images of size $32 \times 32$. The dataset includes 50{,}000 training samples and 10{,}000 test samples. Following common CL practices, we consider three task-splitting configurations: 5, 10, and 20 tasks, each containing an equal number of classes per task. 
For the PixMix data-mixing augmentation method, we use fractal images as the mixing source \citep{hendrycks2022pixmix_ood}, while for Outlier Exposure we used a subset of 300{,}000 samples from the TinyImages dataset containing outliers for training \citep{hendrycks2019oe_ood}. 
\medskip

\myparagraph{Training Details}
We train a ResNet-32 backbone, a lightweight architecture commonly used for CIFAR-like datasets. 
Unless otherwise specified, all models are trained for 200 epochs using SGD with batches of 128 samples, an initial learning rate of 0.1, momentum set to 0.9, and weight decay equal to $5^{-4}$. The learning rate is reduced by a factor of 0.1 after 60, 120 and 170 epochs. Memory-based strategies use a memory buffer of 2,000 samples. 
A few methods require specific adjustments to ensure stable convergence. iCaRL is trained with a weight decay of $10^{-5}$ and a learning rate of 2.0. LwF also uses a weight decay of $10^{-5}$ across all configurations, and when combined with Outlier Exposure, its learning rate is set to 0.01.
We report further details on the hyperparameters selected for each CL and OOD method in the supplementary material.
All experiments are repeated three times using different random seeds and class orders.
\medskip

% \myparagraph{Hardware}
% \todo{Should we describe all the machines we were using? We divided experiments across 4 different workstations :)}
\begin{table*}[h!]
\caption{Continual Learning Results.
For each category, we list the methods from best to worst-performing based on AIA (and ACA in case of ties). We also report, for each dataset, the standard deviation derived from the three different class orderings.}
\label{tab:cl_results}
\centering
\resizebox{0.99\textwidth}{!}{
\begin{tabular}{ll|ccc|ccc|ccc|ccc}
\toprule
& & \multicolumn{3}{c}{\CifarhunFiveT}
& \multicolumn{3}{c}{\CifarhunTenT}
& \multicolumn{3}{c}{\CifarhunTwenT}
& \multicolumn{3}{c}{Average} \\
\cmidrule(lr){3-5} \cmidrule(lr){6-8} \cmidrule(lr){9-11} \cmidrule(lr){12-14} 
& \textbf{Method} & \textbf{ACA}$\uparrow$ & \textbf{AIA}$\uparrow$ & \textbf{AF}$\downarrow$ 
& \textbf{ACA}$\uparrow$ & \textbf{AIA}$\uparrow$ & \textbf{AF}$\downarrow$ 
& \textbf{ACA}$\uparrow$ & \textbf{AIA}$\uparrow$ & \textbf{AF}$\downarrow$
& \textbf{ACA}$\uparrow$ & \textbf{AIA}$\uparrow$ & \textbf{AF}$\downarrow$ \\
% \cmidrule(lr){1-2} \cmidrule(lr){3-5} \cmidrule(lr){6-8} \cmidrule(lr){9-11} \cmidrule(lr){12-14} 
\midrule
(Baseline) & Cumulative & $70.03_{\pm 0.12}$ & $74.61_{\pm 1.23}$ & $4.23_{\pm 0.91}$ & $70.50_{\pm 0.08}$ & $76.30_{\pm 1.18}$ & $4.49_{\pm 0.62}$ & $70.43_{\pm 0.33}$ & $77.12_{\pm 1.19}$ & $4.52_{\pm 0.56}$ & 70.32 & 76.01 & 4.42 \\
\midrule
\multirow{9}{*}{(Memory)} 
& BiC & $49.83_{\pm 0.26}$ & $63.95_{\pm 1.24}$ & $17.53_{\pm 0.45}$ & $45.20_{\pm 0.75}$ & $62.02_{\pm 1.80}$ & $17.77_{\pm 0.28}$ & $41.53_{\pm 0.74}$ & $59.36_{\pm 1.81}$ & $16.43_{\pm 0.20}$ & 45.52 & 61.78 & 17.25 \\
& ER-ACE & $44.33_{\pm 0.52}$ & $59.01_{\pm 1.18}$ & $21.67_{\pm 0.28}$ & $41.43_{\pm 0.66}$ & $57.16_{\pm 1.27}$ & $20.01_{\pm 0.35}$ & $39.47_{\pm 0.31}$ & $55.54_{\pm 1.83}$ & $16.09_{\pm 0.96}$ & 41.74 & 57.24 & 19.26 \\
& DER & $39.63_{\pm 0.19}$ & $58.75_{\pm 1.43}$ & $30.23_{\pm 1.12}$ & $34.90_{\pm 0.82}$ & $56.93_{\pm 1.57}$ & $32.08_{\pm 0.86}$ & $35.03_{\pm 0.61}$ & $55.88_{\pm 1.55}$ & $33.31_{\pm 0.72}$ & 36.52 & 57.19 & 31.87 \\
& ICaRL & $52.00_{\pm 0.36}$ & $62.91_{\pm 1.15}$ & $10.89_{\pm 0.29}$ & $44.33_{\pm 0.81}$ & $57.68_{\pm 2.32}$ & $11.50_{\pm 0.70}$ & $36.70_{\pm 0.67}$ & $49.74_{\pm 2.04}$ & $11.63_{\pm 1.30}$ & 44.34 & 56.77 & 11.34 \\
& IL2M & $36.80_{\pm 0.37}$ & $56.00_{\pm 1.47}$ & $35.28_{\pm 1.15}$ & $34.10_{\pm 0.50}$ & $54.47_{\pm 1.64}$ & $37.26_{\pm 0.63}$ & $33.93_{\pm 0.99}$ & $53.68_{\pm 1.46}$ & $37.83_{\pm 0.54}$ & 34.94 & 54.72 & 36.79 \\
& Replay & $36.37_{\pm 0.12}$ & $55.94_{\pm 1.36}$ & $35.46_{\pm 1.50}$ & $33.67_{\pm 0.98}$ & $54.40_{\pm 1.40}$ & $37.14_{\pm 0.94}$ & $33.67_{\pm 0.65}$ & $53.78_{\pm 1.56}$ & $37.54_{\pm 0.91}$ & 34.57 & 54.71 & 36.71 \\
& RAR & $31.33_{\pm 3.40}$ & $50.09_{\pm 3.61}$ & $23.77_{\pm 1.28}$ & $30.70_{\pm 2.26}$ & $50.60_{\pm 1.28}$ & $23.81_{\pm 0.82}$ & $27.37_{\pm 0.31}$ & $48.20_{\pm 1.15}$ & $26.30_{\pm 0.76}$ & 29.80 & 49.63 & 24.63 \\
& MIR & $32.57_{\pm 5.18}$ & $45.59_{\pm 3.06}$ & $24.59_{\pm 1.90}$ & $33.07_{\pm 2.57}$ & $45.24_{\pm 0.88}$ & $17.22_{\pm 7.80}$ & $32.60_{\pm 1.07}$ & $43.72_{\pm 1.52}$ & $8.30_{\pm 0.84}$ & 32.74 & 44.85 & 16.71 \\
& GDumb & $19.10_{\pm 0.36}$ & $31.85_{\pm 1.16}$ & $14.17_{\pm 1.02}$ & $18.03_{\pm 0.26}$ & $35.54_{\pm 1.88}$ & $18.31_{\pm 0.92}$ & $19.53_{\pm 0.58}$ & $37.59_{\pm 1.38}$ & $19.55_{\pm 0.77}$ & 18.89 & 34.99 & 17.34 \\
\midrule
\multirow{5}{*}{(Regularization)} 
& LwF & $34.07_{\pm 2.35}$ & $53.67_{\pm 2.51}$ & $32.31_{\pm 2.84}$ & $25.60_{\pm 4.17}$ & $45.57_{\pm 1.76}$ & $37.85_{\pm 1.37}$ & $20.33_{\pm 0.21}$ & $35.59_{\pm 2.67}$ & $35.42_{\pm 1.43}$ & 26.67 & 44.94 & 35.19 \\
& GEM & $23.87_{\pm 3.00}$ & $46.81_{\pm 1.44}$ & $44.67_{\pm 2.28}$ & $2.93_{\pm 0.41}$ & $33.49_{\pm 3.71}$ & $51.12_{\pm 2.28}$ & $2.80_{\pm 0.43}$ & $16.77_{\pm 0.49}$ & $35.65_{\pm 4.19}$ & 9.87 & 32.36 & 43.81 \\
& SI & $17.37_{\pm 0.40}$ & $38.01_{\pm 0.56}$ & $66.03_{\pm 1.32}$ & $9.13_{\pm 0.25}$ & $25.65_{\pm 0.59}$ & $78.44_{\pm 1.91}$ & $4.70_{\pm 0.22}$ & $16.40_{\pm 0.27}$ & $85.82_{\pm 0.96}$ & 10.40 & 26.68 & 76.76 \\
& EWC & $17.27_{\pm 0.48}$ & $38.09_{\pm 0.70}$ & $66.23_{\pm 1.53}$ & $9.10_{\pm 0.29}$ & $25.56_{\pm 0.60}$ & $78.35_{\pm 1.92}$ & $4.63_{\pm 0.12}$ & $16.25_{\pm 0.26}$ & $86.23_{\pm 1.14}$ & 10.33 & 26.63 & 76.94 \\
& AGEM & $17.13_{\pm 0.46}$ & $37.72_{\pm 0.78}$ & $65.58_{\pm 1.68}$ & $9.07_{\pm 0.29}$ & $25.49_{\pm 0.55}$ & $78.13_{\pm 1.74}$ & $4.57_{\pm 0.17}$ & $16.21_{\pm 0.25}$ & $86.01_{\pm 1.32}$ & 10.26 & 26.47 & 76.57 \\
\midrule
(Architecture) & DynamicER & $44.90_{\pm 0.51}$ & $61.65_{\pm 1.48}$ & $28.13_{\pm 1.23}$ & $40.37_{\pm 1.18}$ & $59.62_{\pm 1.63}$ & $31.58_{\pm 0.51}$ & $38.50_{\pm 0.14}$ & $58.09_{\pm 1.79}$ & $33.86_{\pm 0.61}$ & 41.26 & 59.79 & 31.19 \\
\bottomrule
\end{tabular}
}
\end{table*}

\section{Experimental Results and Discussion}
\label{sect:exp}
% \mycomment{Budget 10 pages - The goal of this section is to explain the low-level details regarding the experiments and present the experiments that may confirm or not the hypothesis we made before. We should close this section with some guidelines for practitioners (better if in the form of a bullet point list).}
%
In this section, we present the results achieved through our benchmark.
In \autoref{tab:cl_results}, we first report the baseline continual learning performance without any OOD integration. These CL models are then analyzed in combination with post-hoc OOD detectors in \autoref{sect:exp.posthoc} and with training-based OOD adaptations in \autoref{sect:exp.trainingbased}.
\medskip

\myparagraph{CL Baselines}
% \autoref{tab:cl_results} reports the performance of all CL strategies without any training-time OOD method integration. 
% For each CL method, we report its performance in terms of ACA, AIA, and AF, averaged over 3 runs, and evaluated on 5-10-20 task splits. 
To provide an upper bound of the CL performance, we finetune the model using all current and previous data and refer to this method as \textit{Cumulative}. As expected, it achieves the best performance, with ACA and AIA around 70\% and 76\%, respectively, and AF lower than 5\%.
All other methods exhibit performance with a minimum gap of 25\% \wrt the baseline, with constraint-based methods showing the most pronounced degradation.
% Looking at differences across task splits, with the exception of the Cumulative method, which follow a reversed trend, all other CL methods show a progressive decrease in performance as tasks increase. This effect is particularly evident for Constraint-based methods, which start with relatively low performance in the 5-task setting and show a significant drop in ACA together with an increased AF.
% Among the meta-categories, the top-performing methods are memory-based. Looking at the average of the three splits, BiC achieves the highest ACA ($45.52\%$) and AIA ($\sim61.78\%$), demonstrating strong overall discriminative performance. ICaRL, on the other hand, shows the best balance between preserving past knowledge and overall performance, with ACA $44.34\%$ and the lowest AF among replay methods ($11.34\%$). Other replay methods such as ER-ACE and DER show competitive AIA values ($57.24\%$ and $\sim57.19\%$) but suffer from higher forgetting ($\sim19.3\%$ and $\sim31.87\%$), suggesting less stable representations over long sequences of tasks.
Memory-based approaches constitute the strongest category overall. BiC achieves the highest average ACA and AIA across all the three task splits. 
Other approaches of this category, such as ER-ACE and DER, achieve competitive AIA but suffer from higher forgetting.
% Methods relying on simple rehearsal without robust consolidation—such as DER, vanilla Replay, and IL2M—exhibit high forgetting (AF frequently above 30\%) and lower ACA (around 30–35\%), indicating that naive buffer replay alone cannot prevent progressive knowledge degradation. GDumb, as expected, remains a lower bound with ACA below 20\%, serving primarily as a baseline.
Among the constraint-based methods, LwF is the best-performing one, with an AIA of almost 45\%, which is, however, far behind the best memory-based methods. 
% GEM and AGEM have very low ACA (GEM $9.9\%$) and high forgetting, while weight-importance methods (EWC, SI, AGEM in this setup) show the worst behavior, with ACA around 10–10.4\% and catastrophic forgetting (~66–86\% in some splits), demonstrating that weight regularization alone is insufficient to retain usable knowledge across complex datasets like CIFAR-100.
Finally, the architecture-based method DynamicER rank among the top CL approaches (third overall), achieving strong ACA and AIA (41.26\% and 59.79\%, on average). However, they still exhibit non-negligible forgetting around 30\%.

% occupy a middle ground: while they achieve ACA and AIA comparable to mid-tier replay methods, they still suffer from significant forgetting (AF ≈31\%). This suggests that capacity expansion alone does not guarantee robust long-term retention without targeted consolidation or effective replay strategies.

% Show performances in \autoref{tab:cl_results}: ACA and AIA for each CL method. Results averaged over 3 seeds, possibly 3 datasets (CIFAR100-5/10/20 tasks).
% Key messages:
% - we assess the best performing CL methods, category-wise, considering the average across datasets, also averaged across seeds
% - which are the best methods?
% - mention that increasing the number of tasks tends to decrease the CL performance

% In the following, we show that certain CL methods can be successfully combined with both post-hoc and training-based OOD detectors by applying the principles described in \autoref{sect:design} and \autoref{sect:bench}.

\subsection{Combining Post-hoc OOD Detectors with CL} 
\label{sect:exp.posthoc}
\begin{figure*}[ht]
\includegraphics[width=0.9\textwidth]{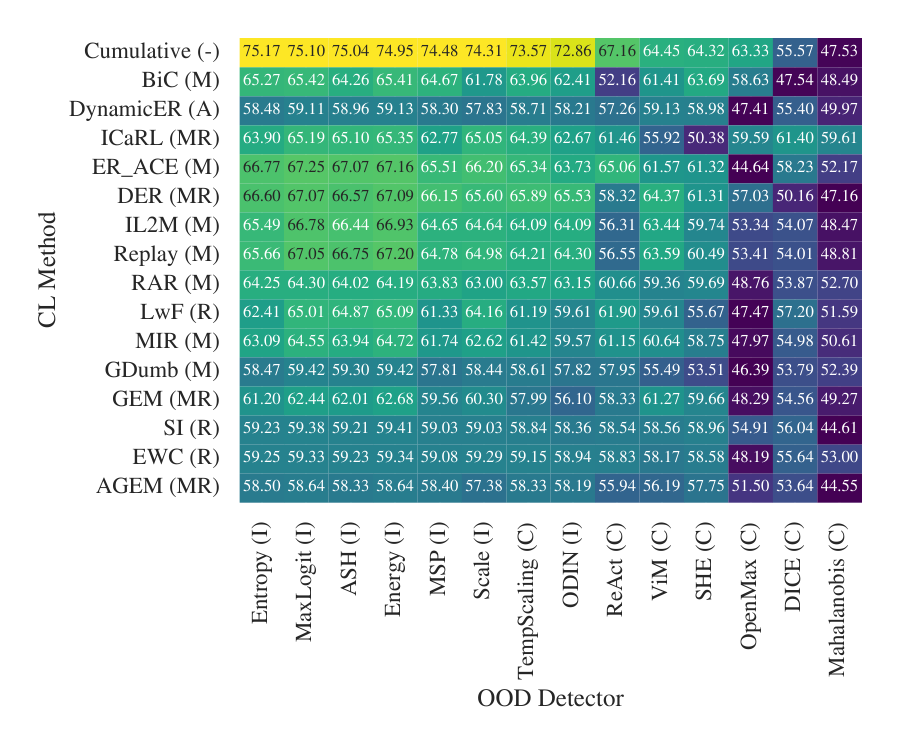}
\caption{AUROC of CL method combined with Post-hoc OOD detectors, measured on CIFAR100-10T. For each CL method and OOD detector, we indicate in brackets its category, as listed in \autoref{tab:supported_methods}. \textit{Rows:} the CL methods are ordered based on AIA, ACA, and AF. \textit{Columns:} the post-hoc OOD detectors ordered based on the AUROC on the Cumulative method.}
\label{fig:hm_posthoc}
\end{figure*}

In \autoref{fig:hm_posthoc}, we show the average AUROC computed over three different seeds, across all tasks from CIFAR100-10T, obtained by applying post-hoc OOD detectors trained on CL methods introduced in \autoref{tab:cl_results}. 
% All detectors are applied after a CL model has completed the training phase on a task. 
% Calibration-based detectors are updated using the data of the current task, and are updated sequentially at every new tasks.
%
Memory-based approaches generally achieve higher AUROC values than regularization-only strategies. In particular, ER-ACE, RAR, and MIR tend to yield relatively strong and consistent results across most detectors, while methods such as LwF, EWC, and SI exhibit lower performance throughout.
Regarding the detectors, the inference-based techniques, MSP, MaxLogit, Energy, and Entropy, consistently achieve the most reliable performance across the majority of CL methods. They are algorithmically and logistically simpler techniques. More complex detectors, such as OpenMax and Mahalanobis, display substantially lower scores in almost all settings, suggesting a higher sensitivity to the representational shifts induced by continual learning.

% \begin{figure*}[ht]
% \includegraphics[width=0.9\linewidth]{posthoc-AvgFPR95TPR-CIFAR100-10T-NOB.pdf}
% \caption{FPR at 95\% TPR in CL method combined with Post-hoc OOD detectors, measured on CIFAR100-10T. For each CL method and OOD detector, we indicate in brackets its category, as listed in \autoref{tab:supported_methods}. \textit{Rows:} the CL methods are ordered based on AIA, ACA, and AF. \textit{Columns:} the post-hoc OOD detectors ordered based on the AUROC on the Cumulative method.}
% \label{fig:hm_posthoc_fpr}
% \end{figure*}

% \begin{figure*}[t]
% \centering
% \includegraphics[width=0.99\textwidth]{AIA_vs_AUC-CIFAR100-5T.pdf}
% \caption{Trade-off between AIA and AUROC for different CL methods on the CIFAR100-5T dataset, where the reported AUROC for each CL method corresponds to the best-performing OOD detector among those evaluated.}
% \label{fig:scatter_OOD_meth_5}
% \end{figure*}

\begin{figure*}[t]
\centering
\includegraphics[width=0.8\textwidth]{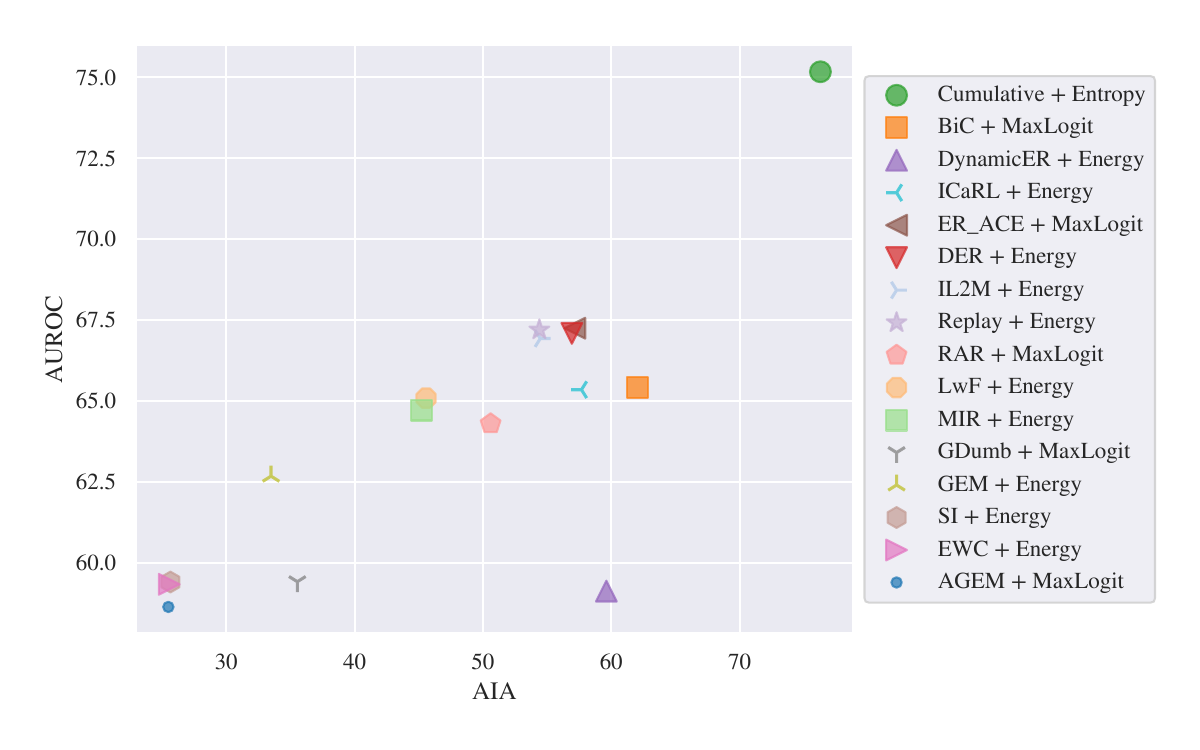}
\caption{Trade-off between AIA and AUROC for different CL methods on the CIFAR100-10T dataset, where the reported AUROC for each CL method corresponds to the best-performing OOD detector among those evaluated.}
\label{fig:scatter_OOD_meth}
\end{figure*}

% \begin{figure*}[t]
% \centering
% \includegraphics[width=0.99\textwidth]{AIA_vs_AUC-CIFAR100-20T.pdf}
% \caption{Trade-off between AIA and AUROC for different CL methods on the CIFAR100-20T dataset, where the reported AUROC for each CL method corresponds to the best-performing OOD detector among those evaluated.}
% \label{fig:scatter_OOD_meth_20}
% \end{figure*}

\myparagraph{Does the best CL and the best detector mean the best combination?}
A first intuitive expectation is that selecting the best CL strategy (in terms of IND accuracy) and the best OOD detector (in terms of AUROC under the Cumulative method) should yield the best overall CL+OOD combination. However, this implicitly assumes that the CL strategy and the OOD detector contribute independently to the final performance. However, \autoref{fig:hm_posthoc} provides direct evidence that this assumption does not hold: interactions between CL training dynamics and OOD scoring can cause combinations built from individually suboptimal components to outperform those based on the best standalone methods.

Although higher accuracy often correlates with better OOD performance in standard static settings, as already noted in \cite{vaze2022openset, miao2026opencil}, this relationship does not hold strictly in a continual learning context.

For instance, LwF, despite achieving only 45\% AIA, attains a higher AUROC than DynamicER, which is otherwise among the top-performing CL methods. On the detector side, the seemingly worst-performing method, \ie Mahalanobis, achieves its best performance when combined with iCaRL, surpassing even the Cumulative method by 12\%. This suggests that training strategies based on class templates may offer an advantage to distance-based OOD detectors.

\autoref{fig:scatter_OOD_meth} further illustrates the relationship between CL and OOD by computing AIA on x-axis and AUROC on y-axis for the best performing combinations on CIFAR100-10T.
% AIA (relevant metric of CL) and the AUROC (relevant metric of OOD) obtained by the best detector on CIFAR100-10T. 
From this analysis, it emerges that the most effective CL+OOD combinations involve ER-ACE, DER, Replay, IL2M, iCaRL, and BiC, particularly when paired with inference-time detectors such as Energy and MaxLogit, as discussed above.

% \begin{figure*}[ht]
% \includegraphics[width=0.99\linewidth]{bestmethods-AvgAUROC-CIFAR100-10T-NOB.pdf}
% % \includegraphics[width=0.45\linewidth]{bestmethods-AvgFPR95TPR-CIFAR100-10T-NOB.pdf}
% \caption{AUROC of CL method combined with Post-hoc OOD detectors, measured on CIFAR100-10T. \textit{Rows:} the CL methods from \autoref{tab:cl_results}, with their category in brackets from \autoref{tab:supported_methods}. \textit{Columns:} the Post-hoc OOD methods, split into inference-time (left) and calibration-time (right).}
% \caption{Average AUROC of each CL method combined with an inference-based OOD detector. The results are averaged over three random class orders. The CL methods are ordered based on AIA, ACA and AF, while OOD detectors are ordered based on the AUROC on the Cumulative method.}
% \label{fig:best_combinations}
% \end{figure*}

\subsection{Combining Training-time OOD Detection with CL}
\label{sect:exp.trainingbased}
In \autoref{tab:ood_aware_cl}, we investigate how training-based OOD strategies interact with different CL methods. The results reveal strongly heterogeneous behaviors: while a few combinations lead to consistent improvements across both CL and OOD metrics \wrt their standard CL training counterparts, many others exhibit pronounced trade-offs, typically improving OOD detection at the expense of accuracy and forgetting.

In particular, LogitNorm applies a constraint on the scale of the logits, effectively reducing reliance on magnitude and emphasizing angular separation.
Under this setting, simple Replay consistently achieves gains of approximately 2–3\% across both CL and OOD metrics. 
In contrast, BiC, iCaRL, and LwF experience noticeable degradations in AIA, with BiC and iCaRL being the most severely affected; for BiC, the modest AUROC improvement does not compensate for the substantial accuracy loss.
The remaining CL strategies, namely ER-ACE and DynamicER, exhibit only marginal performance variations under LogitNorm.
Outlier Exposure (OE) explicitly enforces separation between in-distribution and out-of-distribution samples during each training session.
In this case, Replay and DynamicER benefit across all metrics, although Replay shows a minor drop in AIA in the 10- and 20-task configurations.
All other methods suffer severe degradations across metrics, with BiC being the sole exception, trading these drops for an AUROC improvement of approximately 7\%.
PixMix perturbs the input space rather than directly constraining the logit or feature geometry, resulting in a softer and more CL-compatible form of OOD regularization, which yields more balanced and favorable outcomes overall. 
Notably, all methods benefit from this integration, particularly ICaRL, which performs poorly under LogitNorm and OE but here achieves gains of approximately 7\% in AIA and 5\% in AUROC in the 20-task configuration.
ER-ACE is the only exception, showing a slight decrease in CL performance, accompanied by only negligible AUROC improvements in the 5- and 10-task configurations.

Overall, unlike the setting without training-time OOD integration, DynamicER consistently emerges as the top-performing method in terms of AIA, while Replay, the simplest memory-based strategy, ranks first in terms of AUROC.
This indicates that these two methods are particularly effective at exploiting training-time OOD mechanisms in synergy with continual learning dynamics.

\begin{table*}[t]
\centering
\resizebox{0.9\textwidth}{!}{
\begin{tabular}{ll|cc|cc|cc}
\toprule
\multicolumn{2}{c}{} & \multicolumn{2}{c}{\textit{CIFAR100-5T}} & \multicolumn{2}{c}{\textit{CIFAR100-10T}} & \multicolumn{2}{c}{\textit{CIFAR100-20T}} \\
\cmidrule(lr){3-4} \cmidrule(lr){5-6} \cmidrule(lr){7-8}
 & \textbf{CL Method} & \textbf{AIA}$\uparrow$ & \textbf{AUROC}$\uparrow$ & \textbf{AIA}$\uparrow$ & \textbf{AUROC}$\uparrow$ & \textbf{AIA}$\uparrow$ & \textbf{AUROC}$\uparrow$ \\
\midrule
\multirow{7}{*}{\rotatebox[origin=c]{90}{+ LogitNorm}}
& Cumulative & 74.85 (\textcolor{Green}{+0.24}) & 75.97 (\textcolor{Green}{+1.85}) & 76.40 (\textcolor{Green}{+0.10}) & 75.86 (\textcolor{Green}{+1.38}) & 77.21 (\textcolor{Green}{+0.09}) & 75.84 (\textcolor{Green}{+1.42}) \\
\cmidrule{2-8}
& Replay & 58.08 (\textcolor{Green}{+2.14}) & \textbf{70.88} (\textcolor{Green}{+3.64}) & 56.75 (\textcolor{Green}{+2.35}) & \textbf{68.52} (\textcolor{Green}{+3.74}) & 55.41 (\textcolor{Green}{+1.63}) & \textbf{67.09} (\textcolor{Green}{+3.23}) \\
& ER-ACE & \underline{59.13} (\textcolor{Green}{+0.12}) & 69.12 (\textcolor{Green}{+1.01}) & \underline{57.18} (\textcolor{Green}{+0.02}) & 66.09 (\textcolor{Green}{+0.58}) & \underline{55.66} (\textcolor{Green}{+0.12}) & 64.33 (\textcolor{Green}{+0.08}) \\
& BiC & 42.02 (\textcolor{red}{-21.93}) & \underline{69.93} (\textcolor{Green}{+1.94}) & 37.75 (\textcolor{red}{-24.27}) & \underline{66.86} (\textcolor{Green}{+2.19}) & 37.24 (\textcolor{red}{-22.12}) & \underline{65.85} (\textcolor{Green}{+3.20}) \\
& ICaRL & 47.23 (\textcolor{red}{-15.68}) & 66.40 (\textcolor{red}{-1.27}) & 41.16 (\textcolor{red}{-16.52}) & 62.36 (\textcolor{red}{-0.41}) & 32.88 (\textcolor{red}{-16.86}) & 57.53 (\textcolor{Green}{+1.39}) \\
& LwF & 51.14 (\textcolor{red}{-2.53}) & 65.26 (\textcolor{red}{-0.12}) & 42.46 (\textcolor{red}{-3.11}) & 60.55 (\textcolor{red}{-0.78}) & 28.08 (\textcolor{red}{-7.51}) & 54.84 (\textcolor{red}{-1.84}) \\
& DynamicER & \textbf{61.17} (\textcolor{red}{-0.48}) & 65.18 (\textcolor{Green}{+1.26}) & \textbf{59.72} (\textcolor{Green}{+0.10}) & 58.93 (\textcolor{Green}{+0.63}) & \textbf{57.91} (\textcolor{red}{-0.17}) & 54.11 (\textcolor{Green}{+0.11}) \\
\midrule
\multirow{7}{*}{\rotatebox[origin=c]{90}{+ OE}} 
& Cumulative & 70.96 (\textcolor{red}{-3.65}) & 76.69 (\textcolor{Green}{+2.57}) & 72.63 (\textcolor{red}{-3.67}) & 76.89 (\textcolor{Green}{+2.41}) & 74.39 (\textcolor{red}{-2.73}) & 77.58 (\textcolor{Green}{+3.16}) \\
\cmidrule{2-8}
& Replay & \underline{56.75} (\textcolor{Green}{+0.81}) & \textbf{74.12} (\textcolor{Green}{+6.88}) & 54.09 (\textcolor{red}{-0.31}) & \underline{71.67} (\textcolor{Green}{+6.89}) & 52.69 (\textcolor{red}{-1.09}) & \textbf{70.62} (\textcolor{Green}{+6.76}) \\
& ER-ACE & 55.05 (\textcolor{red}{-3.96}) & 69.67 (\textcolor{Green}{+1.56}) & 50.68 (\textcolor{red}{-6.48}) & 66.12 (\textcolor{Green}{+0.61}) & 45.53 (\textcolor{red}{-10.01}) & 63.79 (\textcolor{red}{-0.46}) \\
& BiC & 55.63 (\textcolor{red}{-8.32}) & \underline{74.12} (\textcolor{Green}{+6.13}) & \underline{55.87} (\textcolor{red}{-6.15}) & \textbf{72.09} (\textcolor{Green}{+7.42}) & \underline{53.96} (\textcolor{red}{-5.40}) & \underline{70.33} (\textcolor{Green}{+7.68}) \\
& ICaRL & 25.50 (\textcolor{red}{-37.41}) & 58.24 (\textcolor{red}{-9.43}) & 24.88 (\textcolor{red}{-32.80}) & 55.15 (\textcolor{red}{-7.62}) & 21.54 (\textcolor{red}{-28.20}) & 53.63 (\textcolor{red}{-2.51}) \\
& LwF & 32.67 (\textcolor{red}{-21.00}) & 62.52 (\textcolor{red}{-2.86}) & 26.26 (\textcolor{red}{-19.31}) & 59.38 (\textcolor{red}{-1.95}) & 19.14 (\textcolor{red}{-16.45}) & 57.22 (\textcolor{Green}{+0.54}) \\
& DynamicER & \textbf{67.68} (\textcolor{Green}{+6.03}) & 66.93 (\textcolor{Green}{+3.01}) & \textbf{66.48} (\textcolor{Green}{+6.86}) & 60.83 (\textcolor{Green}{+2.53}) & \textbf{63.96} (\textcolor{Green}{+5.88}) & 55.64 (\textcolor{Green}{+1.64}) \\
\midrule
\multirow{7}{*}{\rotatebox[origin=c]{90}{+ PixMix}} 
& Cumulative & 75.71 (\textcolor{Green}{+1.10}) & 76.37 (\textcolor{Green}{+2.25}) & 77.33 (\textcolor{Green}{+1.03}) & 76.47 (\textcolor{Green}{+1.99}) & 77.61 (\textcolor{Green}{+0.49}) & 76.37 (\textcolor{Green}{+1.95}) \\
\cmidrule{2-8}
& Replay & 59.21 (\textcolor{Green}{+3.27}) & \underline{70.66} (\textcolor{Green}{+3.42}) & 56.47 (\textcolor{Green}{+2.07}) & \textbf{67.95} (\textcolor{Green}{+3.17}) & 54.47 (\textcolor{Green}{+0.69}) & \underline{66.79} (\textcolor{Green}{+2.93}) \\
& ER-ACE & 58.03 (\textcolor{red}{-0.98}) & 68.61 (\textcolor{Green}{+0.50}) & 55.04 (\textcolor{red}{-2.12}) & 65.79 (\textcolor{Green}{+0.28}) & 50.95 (\textcolor{red}{-4.59}) & 63.46 (\textcolor{red}{-0.79}) \\
& BiC & 64.83 (\textcolor{Green}{+0.88}) & \textbf{70.69} (\textcolor{Green}{+2.70}) & \underline{63.18} (\textcolor{Green}{+1.16}) & \underline{67.93} (\textcolor{Green}{+3.26}) & \underline{61.20} (\textcolor{Green}{+1.84}) & \textbf{66.81} (\textcolor{Green}{+4.16}) \\
& ICaRL & \textbf{66.26} (\textcolor{Green}{+3.35}) & 70.53 (\textcolor{Green}{+2.86}) & 61.97 (\textcolor{Green}{+4.29}) & 65.73 (\textcolor{Green}{+2.96}) & 57.36 (\textcolor{Green}{+7.62}) & 61.22 (\textcolor{Green}{+5.08}) \\
& LwF & 56.49 (\textcolor{Green}{+2.82}) & 68.18 (\textcolor{Green}{+2.80}) & 47.30 (\textcolor{Green}{+1.73}) & 64.35 (\textcolor{Green}{+3.02}) & 41.12 (\textcolor{Green}{+5.53}) & 61.29 (\textcolor{Green}{+4.61}) \\
& DynamicER & \underline{65.16} (\textcolor{Green}{+3.51}) & 64.94 (\textcolor{Green}{+1.02}) & \textbf{64.04} (\textcolor{Green}{+4.42}) & 59.93 (\textcolor{Green}{+1.63}) & \textbf{62.91} (\textcolor{Green}{+4.83}) & 54.40 (\textcolor{Green}{+0.40}) \\
\bottomrule
\end{tabular}
}
\caption{The effect of training-time OOD methods on CL strategies. We show both CL and OOD performance (\%), with AIA and AUROC measured for all three task splits (5, 10, 20 tasks). Results are averaged over three different class orders. For each OOD method (\ie LogitNorm, OE, and PixMix), we highlight the \textbf{first} (in bold) and \underline{second} (underlined) best CL method. We also report in brackets the improvement with respect to the strategies without any training-time OOD adaptations. We use the MSP scoring function to measure the OOD metrics.}
\label{tab:ood_aware_cl}
\end{table*}

% Table with 2 levels rows: 1) CL method (the few selected), 2) the training-based OOD methods 
% Columns: ACA, AIA, AUROC, FPR, averaged for 3 seeds, possibly 3 datasets

\begin{figure*}[t]
\includegraphics[width=0.99\linewidth]{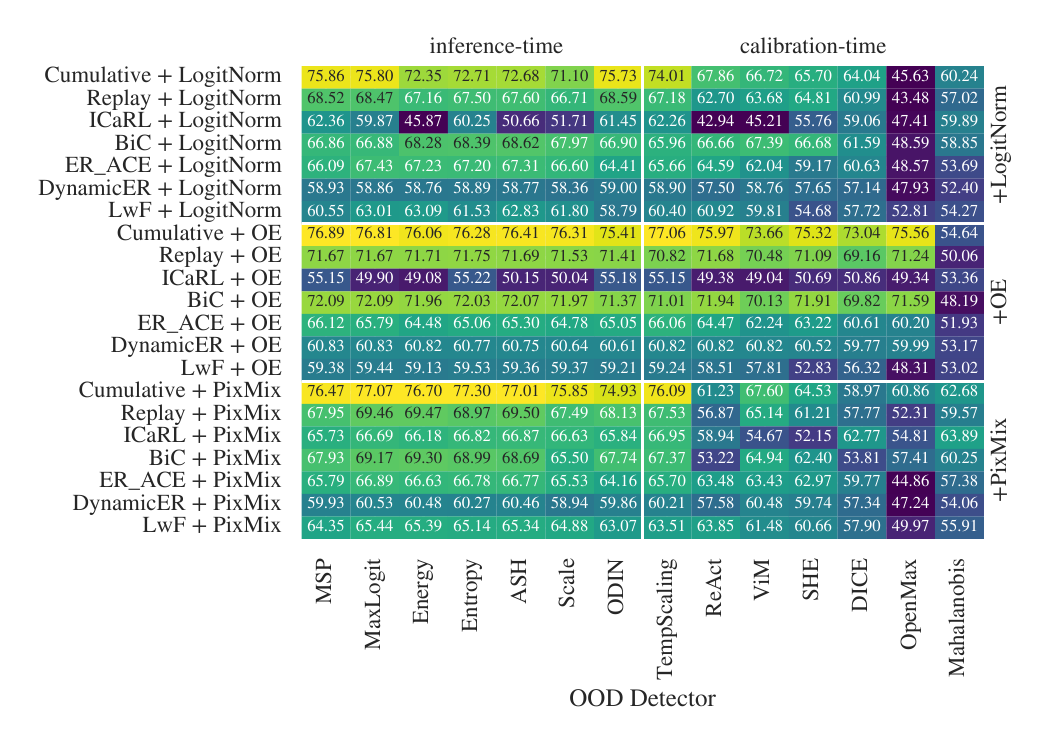}
\caption{AUROC of CL methods combined with training-time OOD integration and post-hoc OOD detectors, measured on CIFAR100-10T.
\textit{Rows:} CL methods combined with LogitNorm, Outlier Exposure (OE), or PixMix.
\textit{Columns:} post-hoc OOD detectors, split into inference-time (left) and calibration-time (right).}
\label{fig:ood_aware_detectors}
\end{figure*}

\medskip
\myparagraph{Further Combining Post-hoc Detectors}
In \autoref{fig:ood_aware_detectors}, we illustrate how each training-based OOD method, once integrated in CL, can yield post-hoc detector to further improve OOD detection performance.
Overall, these results mirror the trends observed in \autoref{fig:hm_posthoc}, where inference-time detectors consistently emerge as the strongest performers, largely independently of the specific training-time OOD integration.
Nevertheless, several method-dependent patterns emerge when analyzing individual training-based strategies.
Under LogitNorm, combining Replay with calibration-time detectors such as ReAct and DICE (both relying on activation pruning) leads to noticeable improvements \wrt the results in \autoref{fig:hm_posthoc}, indicating that logit normalization facilitates their operation.
However, these gains remain insufficient to surpass the best-performing inference-time detectors, such as MSP, MaxLogit, and ODIN.
A similar but more pronounced behavior is observed when using Outlier Exposure.
Here, substantial benefits are confined to two memory-based methods, namely Replay and BiC, for which all detectors achieve AUROC values above 70\%.
As in the LogitNorm case, OE appears to particularly favor activation-pruning detectors like ReAct and DICE, suggesting a consistent interaction between training-time OOD regularization and activation-based calibration mechanisms.
PixMix, instead, yields improvements that are more uniformly distributed across detectors, with inference-time methods again benefiting the most.
Notably, the Mahalanobis distance detector exhibits significant gains—ranging from 5\% to 10\% with respect to standard CL training.

% In \autoref{fig:ood_aware_detectors} we show how the CL methods combined with training-based OOD methods, can be seamlessly combined with different post-hoc detectors by applying the same principles described in ... to obtain further improvements on OOD performance.
% This follows the same trends in \autoref{fig:hm_posthoc}, as inference-time detectors always tend to perform the best independently of the combination employed at training-time. However, some interesting patterns could be noticed. 
% LogitNorm:
% When combining Replay with LogitNorm, ReAct and DICE, two calibration-time detectors relying on activation-pruning, benefit from the LogitNorm integration. Although, not surpassing the best inference-time detectors (MSP, MaxLogit, ODIN).
% %
% Outlier Exposure (OE):
% Here, the advantage is substantial only in two memory-based methods, \ie Replay and BiC, where all the detectors achieves AUROC above 70\%.
% Similar to LogitNorm, OE seem to give an advantage to the activation-pruning methods ReAct and DICE.
% %
% PixMix:
% The improvements are shared over different detectors, mosty inference-time ones.
% Notably, Mahalanobis distance detector improve by 5-10\% \wrt the standard CL training.

% \begin{figure*}[t]
% \includegraphics[width=0.99\linewidth]{heatmap_ood_mod-AvgFPR95TPR-CIFAR100-10T-NOB.pdf}
% \caption{FPR at 95\% TPR in CL methods combined with training-time OOD integration and post-hoc OOD detectors, measured on CIFAR100-10T.
% \textit{Rows:} CL methods combined with LogitNorm, Outlier Exposure (OE), or PixMix.
% \textit{Columns:} post-hoc OOD detectors, split into inference-time (left) and calibration-time (right).}
% \label{fig:ood_aware_detectors_fpr}
% \end{figure*}

% \begin{figure*}[ht]
% \includegraphics[width=0.99\linewidth]{heatmap_ood_mod-diff-AvgAUROC-CIFAR100-10T-NOB.pdf}
% \caption{\textcolor{red}{TO BE REMOVED}}
% \label{fig:ood_aware_detectors}
% \end{figure*}

\subsection{Take-home Messages}
\label{sect:exp.main_messages}
We now summarize the key insights that consistently emerge from our evaluation and highlight practical guidelines for designing CL systems with OOD detection, while also pointing to open challenges that remain insufficiently addressed by current methods.

\begin{enumerate}
\item Having a strong CL model is a necessary but not sufficient condition to obtain strong OOD performance: weaker CL methods may outperform more accurate ones in detection performance if their training dynamics interact more favorably with the chosen OOD detector.

\item Inference-time OOD detectors, especially the simplest ones (\eg Entropy, MaxLogit, Energy), consistently deliver the strongest and most consistent OOD performance across different CL methods, making them the safest default choice when integrating OOD detection into CL pipelines.

% \item Post-hoc (DT) methods behave unpredictably under CL, and their data-dependent nature complicates integration, highlighting the need for a more focused study to model their synergy.
\item Calibration-time detectors (\eg ViM, SHE, OpenMax, Mahalanobis) show method-dependent variability under CL, as their scoring relies on features that continuously change. This suggests the need for more sophisticated strategies to make them compatible, such as a different use of the replay memory (if available) or stronger regularization in feature space. For instance, \cite{kim2022more, gupta2026bufferfree} have shown the potential of calibration-time detectors (in particular Mahalanobis distance) when combined with a particular architecture-based method trained with hard attention masking~\citep{serra2018overcoming} to exclude overwriting of important parameters; this ultimately allows optimal CL performance while maintaining the task-specific representations unmodified, \ie the ideal case for calibration-time detectors to work effectively.

\item Training-based OOD methods markedly improve detection but generally reduce CL accuracy, with the notable exception of a few memory-based strategies and the architecture-based method, where the additional OOD-aware supervision also enhances CL performance.

% \item Current literature lacks dedicated methods that tackle CL and OOD challenges jointly. This can be seen in~\autoref{fig:scatter_OOD_meth} where there is a significant gap in the best performing combined method and the Cumulative baseline.

\end{enumerate}

\section{Related Work}
\label{sect:related}
% \mycomment{Budget 0.5 pages}

Prior work has separately advanced continual learning and out-of-distribution detection, but their joint integration remains underexplored. Classical CL research has focused on mitigating forgetting that is systematically reviewed in recent surveys on CL~\citep{Wang2024tpami_cl, de2021_cl, Zhou202424_cil}. In contrast, the OOD literature focuses on improving separability of in-distribution and out-of-distribution samples as explained in depth in the surveys by~\cite{YangZ2024_ood}, \cite{Lu2024recent_ood}, \cite{Lang2024tmlr_ood}. However, they typically assume static IND scenarios and do not explicitly address the challenges introduced by temporal distribution shifts, common in CL.

A small but growing body of work investigates the joint challenges of CL and OOD detection. \cite{kim2023openworld} theoretically unify novelty detection and class-incremental learning, showing that within-task prediction and OOD detection are necessary and sufficient conditions for good CIL. Both \cite{kim2022more} and \cite{gupta2026bufferfree} perform learning using architecture-based CL methods. While the former uses a buffer to replay OOD data for predicting task-id, the latter combines CL learning with post-hoc OOD detection to perform buffer-free task-id prediction. \cite{Liu2025_clood} provide a hierarchical statistical test to eliminate threshold selection for task-id prediction. All these works focus on a single CL framework and limited OOD methods. They primarily use OOD methods for task-id prediction, rather than to detect wild samples at deployment. \cite{He2022_clood}
formulate OOD detection for unsupervised CL by introducing an output bias correction method to identify whether input samples belong to seen classes without modifying the underlying CL method. Most existing CL-OOD research is limited to theoretical understanding and specific combinations of CL and OOD methods, but lacks a unified empirical evaluation of how diverse OOD methods integrate with major CL paradigms in different stages from learning to inference. \cite{miao2026opencil} represents an early step in this direction by combining quite a small set of CIL methods with several OOD detection techniques. However, the absence of a structured taxonomy and the evaluation over a limited subset of CIL approaches prevent a systematic understanding of how entire categories of CL and OOD methods interact, beyond isolated method-level observations.

Our contribution fills this gap by (i) establishing a unified taxonomy of CL and OOD techniques covering a large landscape of existing works, (ii) introducing a large-scale benchmark suite that systematically evaluates state-of-the-art OOD detectors across representative continual learning scenarios, and (iii) quantifying performance trade-offs and failure modes that are not captured in the current literature.

% Both Continual Learning and Out-of-distribution detection are active fields of research, while solving separate problems, they both work on the same paradigm of adapting models beyond i.i.d. assumption. 

% CL surveys:
% a. \citep{Wang2024tpami_cl}
% b. \citep{de2021_cl}
% c. \cite{Zhou202424_cil}

% OOD surveys:
% a. \citep{YangZ2024_ood}
% b. \citep{Lu2024recent_ood}
% c. \citep{Lang2024tmlr_ood}

% CL-OOD combined work:
% a. \citep{kim2023openworld}
% b. \citep{miao2026opencil}
% c. \citep{kim2022more}
% d. \citep{gupta2026bufferfree}
% e. \citep{He2022_clood}
% f. \citep{Liu2025_clood}

% Here we should provide a picture of the state of the art (related benchmarks+surveys) and highlight their pitfalls and our contribution with respect to them. 

% Battista: ALSO WRT SURVEY PAPERS see these two but I'm sure there are others.

% see TPAMI 2024 \url{https://github.com/LAMDA-CL/CIL_Survey} and \url{https://www.sciencedirect.com/science/article/pii/S0004370224001735} - how are we novel against them? 

\section{Conclusion}
\label{sect:conclusions}
% \mycomment{Budget 0.5 pages}
% Here we should recap: (i) what we have done (ii) the main advancement with respect to the state of the art (iii) the main outcome of our analysis
In this work, we studied the integration of OOD detection into CL systems, a problem that remains largely underexplored despite its relevance for real-world deployment. While CL and OOD detection have been extensively studied in isolation, their broad interaction remains underexplored, as continual updates reshape feature representations and directly affect the assumptions underlying OOD detection methods.

We addressed this gap by proposing (i) a unified taxonomy that jointly organizes CL strategies and OOD detection methods, and (ii) a large-scale benchmark that systematically evaluates their combinations across class-incremental learning scenarios.
Our results reveal consistent interaction patterns between CL dynamics and OOD detection mechanisms. Strong CL performance is necessary but not sufficient for effective OOD detection, and simple inference-time detectors (e.g., Entropy, MaxLogit, Energy) emerge as the most robust choices across a wide range of CL methods. In contrast, calibration-time and training-time OOD approaches exhibit pronounced method-dependent trade-offs, often improving detection at the expense of CL performance, with only a few strategies showing positive synergy.

These findings provide concrete guidance for practitioners designing machine-learning systems that can operate in the open world, and delineate directions for future research. In particular, our study highlights the need for further investigation on representation learning and OOD awareness, building on top of our proposed post-hoc add-on. The proposed taxonomy and benchmark establish a foundation for the systematic development and rigorous evaluation of truly joint CL–OOD approaches.

\section{Acknowledgments}

This work was partially supported by project FAIR (PE00000013) under the NRRP MUR program funded by the EU – NGEU (CUP: J23C24000090007), and by the EU-funded project CoEvolution (grant agreement no. 101168560). This work was conducted while Srishti Gupta was enrolled in the Italian National Doctorate on AI run by Sapienza University of Rome in collaboration with the University of Cagliari.

\section{Statements and Declarations}

\myparagraph{Data Availability} The data supporting the findings of this study are
derived from publicly available datasets, which are well-known in the research community. These datasets have been properly referenced within the paper. For further information on accessing these datasets,
please refer to the references cited within the article.

\myparagraph{Open Access} This article is licensed under a Creative Commons Attribution 4.0 International License, which permits use, sharing, adaptation, distribution and reproduction in any medium or format, as long as you give appropriate credit to the original author(s) and the source, provide a link to the Creative Commons licence, and indicate if changes were made. The images or other third party material in this article are included in the article’s Creative Commons licence,
unless indicated otherwise in a credit line to the material. If material is not included in the article’s Creative Commons licence and your intended use is not permitted by statutory regulation or exceeds the permitted use, you will need to obtain permission directly from the copyright holder. To view a copy of this licence, visit \url{http://creativecommons.org/licenses/by/4.0/}.

\myparagraph{Conflict of Interest} The authors declare that they have no conflict of interest.

% \bibliographystyle{elsarticle-num} 
% \bibliography{bibliography}

\begin{thebibliography}{69}
% BibTex style file: bmc-mathphys.bst (version 2.1), 2014-07-24
\ifx \bisbn   \undefined \def \bisbn  #1{ISBN #1}\fi
\ifx \binits  \undefined \def \binits#1{#1}\fi
\ifx \bauthor  \undefined \def \bauthor#1{#1}\fi
\ifx \batitle  \undefined \def \batitle#1{#1}\fi
\ifx \bjtitle  \undefined \def \bjtitle#1{#1}\fi
\ifx \bvolume  \undefined \def \bvolume#1{\textbf{#1}}\fi
\ifx \byear  \undefined \def \byear#1{#1}\fi
\ifx \bissue  \undefined \def \bissue#1{#1}\fi
\ifx \bfpage  \undefined \def \bfpage#1{#1}\fi
\ifx \blpage  \undefined \def \blpage #1{#1}\fi
\ifx \burl  \undefined \def \burl#1{\textsf{#1}}\fi
\ifx \doiurl  \undefined \def \doiurl#1{\url{https://doi.org/#1}}\fi
\ifx \betal  \undefined \def \betal{\textit{et al.}}\fi
\ifx \binstitute  \undefined \def \binstitute#1{#1}\fi
\ifx \binstitutionaled  \undefined \def \binstitutionaled#1{#1}\fi
\ifx \bctitle  \undefined \def \bctitle#1{#1}\fi
\ifx \beditor  \undefined \def \beditor#1{#1}\fi
\ifx \bpublisher  \undefined \def \bpublisher#1{#1}\fi
\ifx \bbtitle  \undefined \def \bbtitle#1{#1}\fi
\ifx \bedition  \undefined \def \bedition#1{#1}\fi
\ifx \bseriesno  \undefined \def \bseriesno#1{#1}\fi
\ifx \blocation  \undefined \def \blocation#1{#1}\fi
\ifx \bsertitle  \undefined \def \bsertitle#1{#1}\fi
\ifx \bsnm \undefined \def \bsnm#1{#1}\fi
\ifx \bsuffix \undefined \def \bsuffix#1{#1}\fi
\ifx \bparticle \undefined \def \bparticle#1{#1}\fi
\ifx \barticle \undefined \def \barticle#1{#1}\fi
\bibcommenthead
\ifx \bconfdate \undefined \def \bconfdate #1{#1}\fi
\ifx \botherref \undefined \def \botherref #1{#1}\fi
\ifx \url \undefined \def \url#1{\textsf{#1}}\fi
\ifx \bchapter \undefined \def \bchapter#1{#1}\fi
\ifx \bbook \undefined \def \bbook#1{#1}\fi
\ifx \bcomment \undefined \def \bcomment#1{#1}\fi
\ifx \oauthor \undefined \def \oauthor#1{#1}\fi
\ifx \citeauthoryear \undefined \def \citeauthoryear#1{#1}\fi
\ifx \endbibitem  \undefined \def \endbibitem {}\fi
\ifx \bconflocation  \undefined \def \bconflocation#1{#1}\fi
\ifx \arxivurl  \undefined \def \arxivurl#1{\textsf{#1}}\fi
\csname PreBibitemsHook\endcsname

%%% 1
\bibitem[\protect\citeauthoryear{Aljundi et~al.}{2018}]{aljundi2018mas_cl}
\begin{bchapter}
\bauthor{\bsnm{Aljundi}, \binits{R.}},
\bauthor{\bsnm{Babiloni}, \binits{F.}},
\bauthor{\bsnm{Elhoseiny}, \binits{M.}},
\bauthor{\bsnm{Rohrbach}, \binits{M.}},
\bauthor{\bsnm{Tuytelaars}, \binits{T.}}:
\bctitle{Memory aware synapses: Learning what (not) to forget}.
In: \bbtitle{{ECCV} {(3)}}.
\bsertitle{Lecture Notes in Computer Science},
vol. \bseriesno{11207},
pp. \bfpage{144}--\blpage{161}
(\byear{2018})
\end{bchapter}
\endbibitem

%%% 2
\bibitem[\protect\citeauthoryear{Aljundi et~al.}{2019}]{aljundi2019mir_cl}
\begin{botherref}
\oauthor{\bsnm{Aljundi}, \binits{R.}},
\oauthor{\bsnm{Caccia}, \binits{L.}},
\oauthor{\bsnm{Belilovsky}, \binits{E.}},
\oauthor{\bsnm{Caccia}, \binits{M.}},
\oauthor{\bsnm{Lin}, \binits{M.}},
\oauthor{\bsnm{Charlin}, \binits{L.}},
\oauthor{\bsnm{Tuytelaars}, \binits{T.}}:
Online continual learning with maximally interfered retrieval.
CoRR
\textbf{abs/1908.04742}
(2019)
\end{botherref}
\endbibitem

%%% 3
\bibitem[\protect\citeauthoryear{Bendale and Boult}{2016}]{bendale2016openmax_ood}
\begin{bchapter}
\bauthor{\bsnm{Bendale}, \binits{A.}},
\bauthor{\bsnm{Boult}, \binits{T.E.}}:
\bctitle{Towards open set deep networks}.
In: \bbtitle{{CVPR}},
pp. \bfpage{1563}--\blpage{1572}
(\byear{2016})
\end{bchapter}
\endbibitem

%%% 4
\bibitem[\protect\citeauthoryear{Buzzega et~al.}{2020}]{buzzega2020dark_cl}
\begin{bchapter}
\bauthor{\bsnm{Buzzega}, \binits{P.}},
\bauthor{\bsnm{Boschini}, \binits{M.}},
\bauthor{\bsnm{Porrello}, \binits{A.}},
\bauthor{\bsnm{Abati}, \binits{D.}},
\bauthor{\bsnm{Calderara}, \binits{S.}}:
\bctitle{Dark experience for general continual learning: a strong, simple baseline}.
In: \bbtitle{NeurIPS}
(\byear{2020})
\end{bchapter}
\endbibitem

%%% 5
\bibitem[\protect\citeauthoryear{Buzzega et~al.}{2020}]{buzzega2020corr_cl}
\begin{bchapter}
\bauthor{\bsnm{Buzzega}, \binits{P.}},
\bauthor{\bsnm{Boschini}, \binits{M.}},
\bauthor{\bsnm{Porrello}, \binits{A.}},
\bauthor{\bsnm{Calderara}, \binits{S.}}:
\bctitle{Rethinking experience replay: a bag of tricks for continual learning}.
In: \bbtitle{{ICPR}},
pp. \bfpage{2180}--\blpage{2187}
(\byear{2020})
\end{bchapter}
\endbibitem

%%% 6
\bibitem[\protect\citeauthoryear{Belouadah and Popescu}{2019}]{belouadah2019il2m_cl}
\begin{bchapter}
\bauthor{\bsnm{Belouadah}, \binits{E.}},
\bauthor{\bsnm{Popescu}, \binits{A.}}:
\bctitle{{IL2M:} class incremental learning with dual memory}.
In: \bbtitle{{ICCV}},
pp. \bfpage{583}--\blpage{592}
(\byear{2019})
\end{bchapter}
\endbibitem

%%% 7
\bibitem[\protect\citeauthoryear{Caccia et~al.}{2022}]{caccia2021erace_cl}
\begin{bchapter}
\bauthor{\bsnm{Caccia}, \binits{L.}},
\bauthor{\bsnm{Aljundi}, \binits{R.}},
\bauthor{\bsnm{Asadi}, \binits{N.}},
\bauthor{\bsnm{Tuytelaars}, \binits{T.}},
\bauthor{\bsnm{Pineau}, \binits{J.}},
\bauthor{\bsnm{Belilovsky}, \binits{E.}}:
\bctitle{New insights on reducing abrupt representation change in online continual learning}.
In: \bbtitle{{ICLR}}
(\byear{2022})
\end{bchapter}
\endbibitem

%%% 8
\bibitem[\protect\citeauthoryear{Chaudhry et~al.}{2018}]{chaudhry2018rwalk_cl}
\begin{bchapter}
\bauthor{\bsnm{Chaudhry}, \binits{A.}},
\bauthor{\bsnm{Dokania}, \binits{P.K.}},
\bauthor{\bsnm{Ajanthan}, \binits{T.}},
\bauthor{\bsnm{Torr}, \binits{P.H.S.}}:
\bctitle{Riemannian walk for incremental learning: Understanding forgetting and intransigence}.
In: \bbtitle{{ECCV} {(11)}}.
\bsertitle{Lecture Notes in Computer Science},
vol. \bseriesno{11215},
pp. \bfpage{556}--\blpage{572}
(\byear{2018})
\end{bchapter}
\endbibitem

%%% 9
\bibitem[\protect\citeauthoryear{Chaudhry et~al.}{2019}]{chaudhry2019agem_cl}
\begin{bchapter}
\bauthor{\bsnm{Chaudhry}, \binits{A.}},
\bauthor{\bsnm{Ranzato}, \binits{M.}},
\bauthor{\bsnm{Rohrbach}, \binits{M.}},
\bauthor{\bsnm{Elhoseiny}, \binits{M.}}:
\bctitle{Efficient lifelong learning with {A-GEM}}.
In: \bbtitle{{ICLR} (Poster)}
(\byear{2019})
\end{bchapter}
\endbibitem

%%% 10
\bibitem[\protect\citeauthoryear{Djurisic et~al.}{2023}]{djurisic2023ash_ood}
\begin{bchapter}
\bauthor{\bsnm{Djurisic}, \binits{A.}},
\bauthor{\bsnm{Bozanic}, \binits{N.}},
\bauthor{\bsnm{Ashok}, \binits{A.}},
\bauthor{\bsnm{Liu}, \binits{R.}}:
\bctitle{Extremely simple activation shaping for out-of-distribution detection}.
In: \bbtitle{{ICLR}}
(\byear{2023})
\end{bchapter}
\endbibitem

%%% 11
\bibitem[\protect\citeauthoryear{De~Lange et~al.}{2021}]{de2021_cl}
\begin{barticle}
\bauthor{\bsnm{De~Lange}, \binits{M.}},
\bauthor{\bsnm{Aljundi}, \binits{R.}},
\bauthor{\bsnm{Masana}, \binits{M.}},
\bauthor{\bsnm{Parisot}, \binits{S.}},
\bauthor{\bsnm{Jia}, \binits{X.}},
\bauthor{\bsnm{Leonardis}, \binits{A.}},
\bauthor{\bsnm{Slabaugh}, \binits{G.}},
\bauthor{\bsnm{Tuytelaars}, \binits{T.}}:
\batitle{A continual learning survey: Defying forgetting in classification tasks}.
\bjtitle{IEEE transactions on pattern analysis and machine intelligence}
\bvolume{44}(\bissue{7}),
\bfpage{3366}--\blpage{3385}
(\byear{2021})
\end{barticle}
\endbibitem

%%% 12
\bibitem[\protect\citeauthoryear{Ding et~al.}{2025}]{ding2025enhancing_ood}
\begin{botherref}
\oauthor{\bsnm{Ding}, \binits{Y.}},
\oauthor{\bsnm{Liu}, \binits{X.}},
\oauthor{\bsnm{Unger}, \binits{J.}},
\oauthor{\bsnm{Eilertsen}, \binits{G.}}:
Enhancing out-of-distribution detection with extended logit normalization.
arXiv preprint arXiv:2504.11434
(2025)
\end{botherref}
\endbibitem

%%% 13
\bibitem[\protect\citeauthoryear{Fernando et~al.}{2017}]{fernando2017pathnet_cl}
\begin{botherref}
\oauthor{\bsnm{Fernando}, \binits{C.}},
\oauthor{\bsnm{Banarse}, \binits{D.}},
\oauthor{\bsnm{Blundell}, \binits{C.}},
\oauthor{\bsnm{Zwols}, \binits{Y.}},
\oauthor{\bsnm{Ha}, \binits{D.}},
\oauthor{\bsnm{Rusu}, \binits{A.A.}},
\oauthor{\bsnm{Pritzel}, \binits{A.}},
\oauthor{\bsnm{Wierstra}, \binits{D.}}:
Pathnet: Evolution channels gradient descent in super neural networks.
CoRR
\textbf{abs/1701.08734}
(2017)
\end{botherref}
\endbibitem

%%% 14
\bibitem[\protect\citeauthoryear{Gupta et~al.}{2026}]{gupta2026bufferfree}
\begin{barticle}
\bauthor{\bsnm{Gupta}, \binits{S.}},
\bauthor{\bsnm{Angioni}, \binits{D.}},
\bauthor{\bsnm{Pintor}, \binits{M.}},
\bauthor{\bsnm{Demontis}, \binits{A.}},
\bauthor{\bsnm{Sch{\"{o}}nherr}, \binits{L.}},
\bauthor{\bsnm{Roli}, \binits{F.}},
\bauthor{\bsnm{Biggio}, \binits{B.}}:
\batitle{Buffer-free class-incremental learning with out-of-distribution detection}.
\bjtitle{Pattern Recognit.}
\bvolume{172},
\bfpage{112441}
(\byear{2026})
\end{barticle}
\endbibitem

%%% 15
\bibitem[\protect\citeauthoryear{Guo et~al.}{2017}]{guo2017tempscale_ood}
\begin{bchapter}
\bauthor{\bsnm{Guo}, \binits{C.}},
\bauthor{\bsnm{Pleiss}, \binits{G.}},
\bauthor{\bsnm{Sun}, \binits{Y.}},
\bauthor{\bsnm{Weinberger}, \binits{K.Q.}}:
\bctitle{On calibration of modern neural networks}.
In: \bbtitle{{ICML}}.
\bsertitle{Proceedings of Machine Learning Research},
vol. \bseriesno{70},
pp. \bfpage{1321}--\blpage{1330}
(\byear{2017})
\end{bchapter}
\endbibitem

%%% 16
\bibitem[\protect\citeauthoryear{Hendrycks et~al.}{2022}]{hendrycks2022maxlogit_ood}
\begin{bchapter}
\bauthor{\bsnm{Hendrycks}, \binits{D.}},
\bauthor{\bsnm{Basart}, \binits{S.}},
\bauthor{\bsnm{Mazeika}, \binits{M.}},
\bauthor{\bsnm{Zou}, \binits{A.}},
\bauthor{\bsnm{Kwon}, \binits{J.}},
\bauthor{\bsnm{Mostajabi}, \binits{M.}},
\bauthor{\bsnm{Steinhardt}, \binits{J.}},
\bauthor{\bsnm{Song}, \binits{D.}}:
\bctitle{Scaling out-of-distribution detection for real-world settings}.
In: \bbtitle{{ICML}}.
\bsertitle{Proceedings of Machine Learning Research},
vol. \bseriesno{162},
pp. \bfpage{8759}--\blpage{8773}
(\byear{2022})
\end{bchapter}
\endbibitem

%%% 17
\bibitem[\protect\citeauthoryear{Hendrycks and Gimpel}{2017}]{hendrycks2017msp_ood}
\begin{bchapter}
\bauthor{\bsnm{Hendrycks}, \binits{D.}},
\bauthor{\bsnm{Gimpel}, \binits{K.}}:
\bctitle{A baseline for detecting misclassified and out-of-distribution examples in neural networks}.
In: \bbtitle{International Conference on Learning Representations}
(\byear{2017}).
\burl{https://openreview.net/forum?id=Hkg4TI9xl}
\end{bchapter}
\endbibitem

%%% 18
\bibitem[\protect\citeauthoryear{Hu et~al.}{2018}]{Hu2018OvercomingCF}
\begin{bchapter}
\bauthor{\bsnm{Hu}, \binits{W.}},
\bauthor{\bsnm{Lin}, \binits{Z.}},
\bauthor{\bsnm{Liu}, \binits{B.}},
\bauthor{\bsnm{Tao}, \binits{C.}},
\bauthor{\bsnm{Tao}, \binits{Z.}},
\bauthor{\bsnm{Ma}, \binits{J.}},
\bauthor{\bsnm{Zhao}, \binits{D.}},
\bauthor{\bsnm{Yan}, \binits{R.}}:
\bctitle{Overcoming catastrophic forgetting for continual learning via model adaptation}.
In: \bbtitle{International Conference on Learning Representations}
(\byear{2018}).
\burl{https://api.semanticscholar.org/CorpusID:108297547}
\end{bchapter}
\endbibitem

%%% 19
\bibitem[\protect\citeauthoryear{Hendrycks et~al.}{2020}]{hendrycks2020augmix_ood}
\begin{bchapter}
\bauthor{\bsnm{Hendrycks}, \binits{D.}},
\bauthor{\bsnm{Mu}, \binits{N.}},
\bauthor{\bsnm{Cubuk}, \binits{E.D.}},
\bauthor{\bsnm{Zoph}, \binits{B.}},
\bauthor{\bsnm{Gilmer}, \binits{J.}},
\bauthor{\bsnm{Lakshminarayanan}, \binits{B.}}:
\bctitle{Augmix: {A} simple data processing method to improve robustness and uncertainty}.
In: \bbtitle{{ICLR}}
(\byear{2020})
\end{bchapter}
\endbibitem

%%% 20
\bibitem[\protect\citeauthoryear{Hendrycks et~al.}{2019}]{hendrycks2019oe_ood}
\begin{bchapter}
\bauthor{\bsnm{Hendrycks}, \binits{D.}},
\bauthor{\bsnm{Mazeika}, \binits{M.}},
\bauthor{\bsnm{Dietterich}, \binits{T.G.}}:
\bctitle{Deep anomaly detection with outlier exposure}.
In: \bbtitle{{ICLR} (Poster)}
(\byear{2019})
\end{bchapter}
\endbibitem

%%% 21
\bibitem[\protect\citeauthoryear{Hinton et~al.}{2015}]{hinton2015kdCL}
\begin{botherref}
\oauthor{\bsnm{Hinton}, \binits{G.E.}},
\oauthor{\bsnm{Vinyals}, \binits{O.}},
\oauthor{\bsnm{Dean}, \binits{J.}}:
Distilling the knowledge in a neural network.
CoRR
\textbf{abs/1503.02531}
(2015)
\end{botherref}
\endbibitem

%%% 22
\bibitem[\protect\citeauthoryear{He and Zhu}{2022}]{He2022_clood}
\begin{bchapter}
\bauthor{\bsnm{He}, \binits{J.}},
\bauthor{\bsnm{Zhu}, \binits{F.}}:
\bctitle{Out-of-distribution detection in unsupervised continual learning}.
In: \bbtitle{{CVPR} Workshops},
pp. \bfpage{3849}--\blpage{3854}
(\byear{2022})
\end{bchapter}
\endbibitem

%%% 23
\bibitem[\protect\citeauthoryear{Hendrycks et~al.}{2022}]{hendrycks2022pixmix_ood}
\begin{bchapter}
\bauthor{\bsnm{Hendrycks}, \binits{D.}},
\bauthor{\bsnm{Zou}, \binits{A.}},
\bauthor{\bsnm{Mazeika}, \binits{M.}},
\bauthor{\bsnm{Tang}, \binits{L.}},
\bauthor{\bsnm{Li}, \binits{B.}},
\bauthor{\bsnm{Song}, \binits{D.}},
\bauthor{\bsnm{Steinhardt}, \binits{J.}}:
\bctitle{Pixmix: Dreamlike pictures comprehensively improve safety measures}.
In: \bbtitle{{CVPR}},
pp. \bfpage{16762}--\blpage{16771}
(\byear{2022})
\end{bchapter}
\endbibitem

%%% 24
\bibitem[\protect\citeauthoryear{Jiang et~al.}{2021}]{jiang2021ibdrrCL}
\begin{bchapter}
\bauthor{\bsnm{Jiang}, \binits{J.}},
\bauthor{\bsnm{Cetin}, \binits{E.}},
\bauthor{\bsnm{{\c{C}}eliktutan}, \binits{O.}}:
\bctitle{{IB-DRR} - incremental learning with information-back discrete representation replay}.
In: \bbtitle{{CVPR} Workshops},
pp. \bfpage{3533}--\blpage{3542}
(\byear{2021})
\end{bchapter}
\endbibitem

%%% 25
\bibitem[\protect\citeauthoryear{Kirchheim et~al.}{2022}]{kirchheim2022pytorchood}
\begin{bchapter}
\bauthor{\bsnm{Kirchheim}, \binits{K.}},
\bauthor{\bsnm{Filax}, \binits{M.}},
\bauthor{\bsnm{Ortmeier}, \binits{F.}}:
\bctitle{Pytorch-ood: {A} library for out-of-distribution detection based on pytorch}.
In: \bbtitle{{CVPR} Workshops},
pp. \bfpage{4350}--\blpage{4359}
(\byear{2022})
\end{bchapter}
\endbibitem

%%% 26
\bibitem[\protect\citeauthoryear{Krizhevsky et~al.}{2009}]{krizhevsky2009learning}
\begin{botherref}
\oauthor{\bsnm{Krizhevsky}, \binits{A.}},
\oauthor{\bsnm{Hinton}, \binits{G.}}, et al.:
Learning multiple layers of features from tiny images
(2009)
\end{botherref}
\endbibitem

%%% 27
\bibitem[\protect\citeauthoryear{Kemker and Kanan}{2018}]{kemker2018fearnetCL}
\begin{bchapter}
\bauthor{\bsnm{Kemker}, \binits{R.}},
\bauthor{\bsnm{Kanan}, \binits{C.}}:
\bctitle{Fearnet: Brain-inspired model for incremental learning}.
In: \bbtitle{{ICLR} (Poster)}
(\byear{2018})
\end{bchapter}
\endbibitem

%%% 28
\bibitem[\protect\citeauthoryear{Kim et~al.}{2022}]{kim2022more}
\begin{bchapter}
\bauthor{\bsnm{Kim}, \binits{G.}},
\bauthor{\bsnm{Liu}, \binits{B.}},
\bauthor{\bsnm{Ke}, \binits{Z.}}:
\bctitle{A multi-head model for continual learning via out-of-distribution replay}.
In: \bbtitle{Conference on Lifelong Learning Agents},
pp. \bfpage{548}--\blpage{563}
(\byear{2022}).
\bcomment{PMLR}
\end{bchapter}
\endbibitem

%%% 29
\bibitem[\protect\citeauthoryear{Kirkpatrick et~al.}{2017}]{Kirkpatrick2017ewc_cl}
\begin{barticle}
\bauthor{\bsnm{Kirkpatrick}, \binits{J.}},
\bauthor{\bsnm{Pascanu}, \binits{R.}},
\bauthor{\bsnm{Rabinowitz}, \binits{N.}},
\bauthor{\bsnm{Veness}, \binits{J.}},
\bauthor{\bsnm{Desjardins}, \binits{G.}},
\bauthor{\bsnm{Rusu}, \binits{A.A.}},
\bauthor{\bsnm{Milan}, \binits{K.}},
\bauthor{\bsnm{Quan}, \binits{J.}},
\bauthor{\bsnm{Ramalho}, \binits{T.}},
\bauthor{\bsnm{Grabska-Barwinska}, \binits{A.}},
\bauthor{\bsnm{Hassabis}, \binits{D.}},
\bauthor{\bsnm{Clopath}, \binits{C.}},
\bauthor{\bsnm{Kumaran}, \binits{D.}},
\bauthor{\bsnm{Hadsell}, \binits{R.}}:
\batitle{Overcoming catastrophic forgetting in neural networks}.
\bjtitle{Proceedings of the National Academy of Sciences}
\bvolume{114}(\bissue{13}),
\bfpage{3521}--\blpage{3526}
(\byear{2017})
\doiurl{10.1073/pnas.1611835114}
\end{barticle}
\endbibitem

%%% 30
\bibitem[\protect\citeauthoryear{Kumari et~al.}{2022}]{kumari2022rar_cl}
\begin{bchapter}
\bauthor{\bsnm{Kumari}, \binits{L.}},
\bauthor{\bsnm{Wang}, \binits{S.}},
\bauthor{\bsnm{Zhou}, \binits{T.}},
\bauthor{\bsnm{Bilmes}, \binits{J.A.}}:
\bctitle{Retrospective adversarial replay for continual learning}.
In: \bbtitle{NeurIPS}
(\byear{2022})
\end{bchapter}
\endbibitem

%%% 31
\bibitem[\protect\citeauthoryear{Kim et~al.}{2025}]{kim2023openworld}
\begin{barticle}
\bauthor{\bsnm{Kim}, \binits{G.}},
\bauthor{\bsnm{Xiao}, \binits{C.}},
\bauthor{\bsnm{Konishi}, \binits{T.}},
\bauthor{\bsnm{Ke}, \binits{Z.}},
\bauthor{\bsnm{Liu}, \binits{B.}}:
\batitle{Open-world continual learning: Unifying novelty detection and continual learning}.
\bjtitle{Artif. Intell.}
\bvolume{338},
\bfpage{104237}
(\byear{2025})
\end{barticle}
\endbibitem

%%% 32
\bibitem[\protect\citeauthoryear{Li and Hoiem}{2017}]{li2017lwf_cl}
\begin{botherref}
\oauthor{\bsnm{Li}, \binits{Z.}},
\oauthor{\bsnm{Hoiem}, \binits{D.}}:
Learning without Forgetting
(2017)
\end{botherref}
\endbibitem

%%% 33
\bibitem[\protect\citeauthoryear{Lee et~al.}{2018}]{lee2018md_ood}
\begin{bchapter}
\bauthor{\bsnm{Lee}, \binits{K.}},
\bauthor{\bsnm{Lee}, \binits{K.}},
\bauthor{\bsnm{Lee}, \binits{H.}},
\bauthor{\bsnm{Shin}, \binits{J.}}:
\bctitle{A simple unified framework for detecting out-of-distribution samples and adversarial attacks}.
In: \bbtitle{NeurIPS},
pp. \bfpage{7167}--\blpage{7177}
(\byear{2018})
\end{bchapter}
\endbibitem

%%% 34
\bibitem[\protect\citeauthoryear{Liang et~al.}{2018}]{liang2018odin_ood}
\begin{bchapter}
\bauthor{\bsnm{Liang}, \binits{S.}},
\bauthor{\bsnm{Li}, \binits{Y.}},
\bauthor{\bsnm{Srikant}, \binits{R.}}:
\bctitle{Enhancing the reliability of out-of-distribution image detection in neural networks}.
In: \bbtitle{{ICLR} (Poster)}
(\byear{2018})
\end{bchapter}
\endbibitem

%%% 35
\bibitem[\protect\citeauthoryear{Lomonaco et~al.}{2021}]{lomonaco2021avalanche}
\begin{bchapter}
\bauthor{\bsnm{Lomonaco}, \binits{V.}},
\bauthor{\bsnm{Pellegrini}, \binits{L.}},
\bauthor{\bsnm{Cossu}, \binits{A.}},
\bauthor{\bsnm{Carta}, \binits{A.}},
\bauthor{\bsnm{Graffieti}, \binits{G.}},
\bauthor{\bsnm{Hayes}, \binits{T.L.}},
\bauthor{\bsnm{Lange}, \binits{M.D.}},
\bauthor{\bsnm{Masana}, \binits{M.}},
\bauthor{\bsnm{Pomponi}, \binits{J.}},
\bauthor{\bsnm{Ven}, \binits{G.M.}},
\bauthor{\bsnm{Mundt}, \binits{M.}},
\bauthor{\bsnm{She}, \binits{Q.}},
\bauthor{\bsnm{Cooper}, \binits{K.W.}},
\bauthor{\bsnm{Forest}, \binits{J.}},
\bauthor{\bsnm{Belouadah}, \binits{E.}},
\bauthor{\bsnm{Calderara}, \binits{S.}},
\bauthor{\bsnm{Parisi}, \binits{G.I.}},
\bauthor{\bsnm{Cuzzolin}, \binits{F.}},
\bauthor{\bsnm{Tolias}, \binits{A.S.}},
\bauthor{\bsnm{Scardapane}, \binits{S.}},
\bauthor{\bsnm{Antiga}, \binits{L.}},
\bauthor{\bsnm{Ahmad}, \binits{S.}},
\bauthor{\bsnm{Popescu}, \binits{A.}},
\bauthor{\bsnm{Kanan}, \binits{C.}},
\bauthor{\bsnm{Weijer}, \binits{J.}},
\bauthor{\bsnm{Tuytelaars}, \binits{T.}},
\bauthor{\bsnm{Bacciu}, \binits{D.}},
\bauthor{\bsnm{Maltoni}, \binits{D.}}:
\bctitle{Avalanche: An end-to-end library for continual learning}.
In: \bbtitle{{CVPR} Workshops},
pp. \bfpage{3600}--\blpage{3610}
(\byear{2021})
\end{bchapter}
\endbibitem

%%% 36
\bibitem[\protect\citeauthoryear{Lopez{-}Paz and Ranzato}{2017}]{lopez2017gem_cl}
\begin{bchapter}
\bauthor{\bsnm{Lopez{-}Paz}, \binits{D.}},
\bauthor{\bsnm{Ranzato}, \binits{M.}}:
\bctitle{Gradient episodic memory for continual learning}.
In: \bbtitle{{NeurIPS}},
pp. \bfpage{6467}--\blpage{6476}
(\byear{2017})
\end{bchapter}
\endbibitem

%%% 37
\bibitem[\protect\citeauthoryear{Lange and Tuytelaars}{2021}]{Lange2021prototype_cl}
\begin{bchapter}
\bauthor{\bsnm{Lange}, \binits{M.D.}},
\bauthor{\bsnm{Tuytelaars}, \binits{T.}}:
\bctitle{Continual prototype evolution: Learning online from non-stationary data streams}.
In: \bbtitle{{ICCV}},
pp. \bfpage{8230}--\blpage{8239}
(\byear{2021})
\end{bchapter}
\endbibitem

%%% 38
\bibitem[\protect\citeauthoryear{Liu et~al.}{2020}]{liu2020energy_ood}
\begin{bchapter}
\bauthor{\bsnm{Liu}, \binits{W.}},
\bauthor{\bsnm{Wang}, \binits{X.}},
\bauthor{\bsnm{Owens}, \binits{J.D.}},
\bauthor{\bsnm{Li}, \binits{Y.}}:
\bctitle{Energy-based out-of-distribution detection}.
In: \bbtitle{NeurIPS}
(\byear{2020})
\end{bchapter}
\endbibitem

%%% 39
\bibitem[\protect\citeauthoryear{Lu et~al.}{2024}]{Lu2024recent_ood}
\begin{botherref}
\oauthor{\bsnm{Lu}, \binits{S.}},
\oauthor{\bsnm{Wang}, \binits{Y.}},
\oauthor{\bsnm{Sheng}, \binits{L.}},
\oauthor{\bsnm{Zheng}, \binits{A.}},
\oauthor{\bsnm{He}, \binits{L.}},
\oauthor{\bsnm{Liang}, \binits{J.}}:
Recent advances in {OOD} detection: Problems and approaches.
CoRR
\textbf{abs/2409.11884}
(2024)
\end{botherref}
\endbibitem

%%% 40
\bibitem[\protect\citeauthoryear{Liu et~al.}{2025}]{Liu2025_clood}
\begin{bchapter}
\bauthor{\bsnm{Liu}, \binits{Y.}},
\bauthor{\bsnm{Zhao}, \binits{W.}},
\bauthor{\bsnm{Guo}, \binits{Y.}}:
\bctitle{{H2ST:} hierarchical two-sample tests for continual out-of-distribution detection}.
In: \bbtitle{{CVPR}},
pp. \bfpage{15413}--\blpage{15423}
(\byear{2025})
\end{bchapter}
\endbibitem

%%% 41
\bibitem[\protect\citeauthoryear{Lang et~al.}{2024}]{Lang2024tmlr_ood}
\begin{botherref}
\oauthor{\bsnm{Lang}, \binits{H.}},
\oauthor{\bsnm{Zheng}, \binits{Y.}},
\oauthor{\bsnm{Li}, \binits{Y.}},
\oauthor{\bsnm{Sun}, \binits{J.}},
\oauthor{\bsnm{Huang}, \binits{F.}},
\oauthor{\bsnm{Li}, \binits{Y.}}:
A survey on out-of-distribution detection in {NLP}.
Trans. Mach. Learn. Res.
\textbf{2024}
(2024)
\end{botherref}
\endbibitem

%%% 42
\bibitem[\protect\citeauthoryear{Mallya et~al.}{2018}]{mallya2018piggyback_cl}
\begin{bchapter}
\bauthor{\bsnm{Mallya}, \binits{A.}},
\bauthor{\bsnm{Davis}, \binits{D.}},
\bauthor{\bsnm{Lazebnik}, \binits{S.}}:
\bctitle{Piggyback: Adapting a single network to multiple tasks by learning to mask weights}.
In: \bbtitle{{ECCV} {(4)}}.
\bsertitle{Lecture Notes in Computer Science},
vol. \bseriesno{11208},
pp. \bfpage{72}--\blpage{88}
(\byear{2018})
\end{bchapter}
\endbibitem

%%% 43
\bibitem[\protect\citeauthoryear{Miao et~al.}{2026}]{miao2026opencil}
\begin{barticle}
\bauthor{\bsnm{Miao}, \binits{W.}},
\bauthor{\bsnm{Pang}, \binits{G.}},
\bauthor{\bsnm{Nguyen}, \binits{T.}},
\bauthor{\bsnm{Fang}, \binits{R.}},
\bauthor{\bsnm{Zheng}, \binits{J.}},
\bauthor{\bsnm{Bai}, \binits{X.}}:
\batitle{Opencil: Benchmarking out-of-distribution detection in class incremental learning}.
\bjtitle{Pattern Recognit.}
\bvolume{171},
\bfpage{112163}
(\byear{2026})
\end{barticle}
\endbibitem

%%% 44
\bibitem[\protect\citeauthoryear{Mac{\^{e}}do et~al.}{2021}]{macedo2021entropy_ood}
\begin{bchapter}
\bauthor{\bsnm{Mac{\^{e}}do}, \binits{D.L.}},
\bauthor{\bsnm{Ren}, \binits{T.I.}},
\bauthor{\bsnm{Zanchettin}, \binits{C.}},
\bauthor{\bsnm{Oliveira}, \binits{A.L.I.}},
\bauthor{\bsnm{Ludermir}, \binits{T.B.}}:
\bctitle{Entropic out-of-distribution detection}.
In: \bbtitle{{IJCNN}},
pp. \bfpage{1}--\blpage{8}
(\byear{2021})
\end{bchapter}
\endbibitem

%%% 45
\bibitem[\protect\citeauthoryear{Petit et~al.}{2023}]{petit2023fetrilCL}
\begin{bchapter}
\bauthor{\bsnm{Petit}, \binits{G.}},
\bauthor{\bsnm{Popescu}, \binits{A.}},
\bauthor{\bsnm{Schindler}, \binits{H.}},
\bauthor{\bsnm{Picard}, \binits{D.}},
\bauthor{\bsnm{Delezoide}, \binits{B.}}:
\bctitle{Fetril: Feature translation for exemplar-free class-incremental learning}.
In: \bbtitle{{WACV}},
pp. \bfpage{3900}--\blpage{3909}
(\byear{2023})
\end{bchapter}
\endbibitem

%%% 46
\bibitem[\protect\citeauthoryear{Prabhu et~al.}{2020}]{prabhu2020gdumb_cl}
\begin{bchapter}
\bauthor{\bsnm{Prabhu}, \binits{A.}},
\bauthor{\bsnm{Torr}, \binits{P.H.S.}},
\bauthor{\bsnm{Dokania}, \binits{P.K.}}:
\bctitle{Gdumb: {A} simple approach that questions our progress in continual learning}.
In: \bbtitle{{ECCV} {(2)}}.
\bsertitle{Lecture Notes in Computer Science},
vol. \bseriesno{12347},
pp. \bfpage{524}--\blpage{540}
(\byear{2020})
\end{bchapter}
\endbibitem

%%% 47
\bibitem[\protect\citeauthoryear{Rolnick et~al.}{2019}]{rolnick2019er_cl}
\begin{botherref}
\oauthor{\bsnm{Rolnick}, \binits{D.}},
\oauthor{\bsnm{Ahuja}, \binits{A.}},
\oauthor{\bsnm{Schwarz}, \binits{J.}},
\oauthor{\bsnm{Lillicrap}, \binits{T.P.}},
\oauthor{\bsnm{Wayne}, \binits{G.}}:
Experience Replay for Continual Learning
(2019)
\end{botherref}
\endbibitem

%%% 48
\bibitem[\protect\citeauthoryear{Rebuffi et~al.}{2017}]{rebuffi2017icarl_cl}
\begin{botherref}
\oauthor{\bsnm{Rebuffi}, \binits{S.-A.}},
\oauthor{\bsnm{Kolesnikov}, \binits{A.}},
\oauthor{\bsnm{Sperl}, \binits{G.}},
\oauthor{\bsnm{Lampert}, \binits{C.H.}}:
iCaRL: Incremental Classifier and Representation Learning
(2017)
\end{botherref}
\endbibitem

%%% 49
\bibitem[\protect\citeauthoryear{Rusu et~al.}{2016}]{rusu2022pnn_cl}
\begin{botherref}
\oauthor{\bsnm{Rusu}, \binits{A.A.}},
\oauthor{\bsnm{Rabinowitz}, \binits{N.C.}},
\oauthor{\bsnm{Desjardins}, \binits{G.}},
\oauthor{\bsnm{Soyer}, \binits{H.}},
\oauthor{\bsnm{Kirkpatrick}, \binits{J.}},
\oauthor{\bsnm{Kavukcuoglu}, \binits{K.}},
\oauthor{\bsnm{Pascanu}, \binits{R.}},
\oauthor{\bsnm{Hadsell}, \binits{R.}}:
Progressive neural networks.
CoRR
\textbf{abs/1606.04671}
(2016)
\end{botherref}
\endbibitem

%%% 50
\bibitem[\protect\citeauthoryear{Sun et~al.}{2021}]{sun2021react_ood}
\begin{bchapter}
\bauthor{\bsnm{Sun}, \binits{Y.}},
\bauthor{\bsnm{Guo}, \binits{C.}},
\bauthor{\bsnm{Li}, \binits{Y.}}:
\bctitle{React: Out-of-distribution detection with rectified activations}.
In: \bbtitle{Annual Conf. on Neural Information Processing Systems 2021},
pp. \bfpage{144}--\blpage{157}
(\byear{2021})
\end{bchapter}
\endbibitem

%%% 51
\bibitem[\protect\citeauthoryear{Sun and Li}{2022}]{sun2022dice_ood}
\begin{bchapter}
\bauthor{\bsnm{Sun}, \binits{Y.}},
\bauthor{\bsnm{Li}, \binits{Y.}}:
\bctitle{{DICE:} leveraging sparsification for out-of-distribution detection}.
In: \bbtitle{{ECCV} {(24)}}.
\bsertitle{Lecture Notes in Computer Science},
vol. \bseriesno{13684},
pp. \bfpage{691}--\blpage{708}
(\byear{2022})
\end{bchapter}
\endbibitem

%%% 52
\bibitem[\protect\citeauthoryear{Shin et~al.}{2017}]{shin2017dgr_cl}
\begin{bchapter}
\bauthor{\bsnm{Shin}, \binits{H.}},
\bauthor{\bsnm{Lee}, \binits{J.K.}},
\bauthor{\bsnm{Kim}, \binits{J.}},
\bauthor{\bsnm{Kim}, \binits{J.}}:
\bctitle{Continual learning with deep generative replay}.
In: \bbtitle{{NeurIPS}},
pp. \bfpage{2990}--\blpage{2999}
(\byear{2017})
\end{bchapter}
\endbibitem

%%% 53
\bibitem[\protect\citeauthoryear{Sun et~al.}{2022}]{sun2022knn_ood}
\begin{bchapter}
\bauthor{\bsnm{Sun}, \binits{Y.}},
\bauthor{\bsnm{Ming}, \binits{Y.}},
\bauthor{\bsnm{Zhu}, \binits{X.}},
\bauthor{\bsnm{Li}, \binits{Y.}}:
\bctitle{Out-of-distribution detection with deep nearest neighbors}.
In: \bbtitle{{ICML}}.
\bsertitle{Proceedings of Machine Learning Research},
vol. \bseriesno{162},
pp. \bfpage{20827}--\blpage{20840}
(\byear{2022})
\end{bchapter}
\endbibitem

%%% 54
\bibitem[\protect\citeauthoryear{Serra et~al.}{2018}]{serra2018overcoming}
\begin{bchapter}
\bauthor{\bsnm{Serra}, \binits{J.}},
\bauthor{\bsnm{Suris}, \binits{D.}},
\bauthor{\bsnm{Miron}, \binits{M.}},
\bauthor{\bsnm{Karatzoglou}, \binits{A.}}:
\bctitle{Overcoming catastrophic forgetting with hard attention to the task}.
In: \bbtitle{International Conference on Machine Learning},
pp. \bfpage{4548}--\blpage{4557}
(\byear{2018}).
\bcomment{PMLR}
\end{bchapter}
\endbibitem

%%% 55
\bibitem[\protect\citeauthoryear{Vaze et~al.}{2022}]{vaze2022openset}
\begin{bchapter}
\bauthor{\bsnm{Vaze}, \binits{S.}},
\bauthor{\bsnm{Han}, \binits{K.}},
\bauthor{\bsnm{Vedaldi}, \binits{A.}},
\bauthor{\bsnm{Zisserman}, \binits{A.}}:
\bctitle{Open-set recognition: {A} good closed-set classifier is all you need}.
In: \bbtitle{{ICLR}}
(\byear{2022})
\end{bchapter}
\endbibitem

%%% 56
\bibitem[\protect\citeauthoryear{Wu et~al.}{2019a}]{wu2019bic_cl}
\begin{bchapter}
\bauthor{\bsnm{Wu}, \binits{Y.}},
\bauthor{\bsnm{Chen}, \binits{Y.}},
\bauthor{\bsnm{Wang}, \binits{L.}},
\bauthor{\bsnm{Ye}, \binits{Y.}},
\bauthor{\bsnm{Liu}, \binits{Z.}},
\bauthor{\bsnm{Guo}, \binits{Y.}},
\bauthor{\bsnm{Fu}, \binits{Y.}}:
\bctitle{Large scale incremental learning}.
In: \bbtitle{{CVPR}},
pp. \bfpage{374}--\blpage{382}
(\byear{2019})
\end{bchapter}
\endbibitem

%%% 57
\bibitem[\protect\citeauthoryear{Wu et~al.}{2019b}]{Wu19bic_cl}
\begin{bchapter}
\bauthor{\bsnm{Wu}, \binits{Y.}},
\bauthor{\bsnm{Chen}, \binits{Y.}},
\bauthor{\bsnm{Wang}, \binits{L.}},
\bauthor{\bsnm{Ye}, \binits{Y.}},
\bauthor{\bsnm{Liu}, \binits{Z.}},
\bauthor{\bsnm{Guo}, \binits{Y.}},
\bauthor{\bsnm{Fu}, \binits{Y.}}:
\bctitle{Large scale incremental learning}.
In: \bbtitle{{CVPR}},
pp. \bfpage{374}--\blpage{382}
(\byear{2019})
\end{bchapter}
\endbibitem

%%% 58
\bibitem[\protect\citeauthoryear{Wang et~al.}{2022}]{wang2022vim_ood}
\begin{bchapter}
\bauthor{\bsnm{Wang}, \binits{H.}},
\bauthor{\bsnm{Li}, \binits{Z.}},
\bauthor{\bsnm{Feng}, \binits{L.}},
\bauthor{\bsnm{Zhang}, \binits{W.}}:
\bctitle{Vim: Out-of-distribution with virtual-logit matching}.
In: \bbtitle{{CVPR}},
pp. \bfpage{4911}--\blpage{4920}
(\byear{2022})
\end{bchapter}
\endbibitem

%%% 59
\bibitem[\protect\citeauthoryear{Wang et~al.}{2021}]{Wang2021adam-nscl_cl}
\begin{bchapter}
\bauthor{\bsnm{Wang}, \binits{S.}},
\bauthor{\bsnm{Li}, \binits{X.}},
\bauthor{\bsnm{Sun}, \binits{J.}},
\bauthor{\bsnm{Xu}, \binits{Z.}}:
\bctitle{Training networks in null space of feature covariance for continual learning}.
In: \bbtitle{{CVPR}},
pp. \bfpage{184}--\blpage{193}
(\byear{2021})
\end{bchapter}
\endbibitem

%%% 60
\bibitem[\protect\citeauthoryear{Wei et~al.}{2022}]{wei2022logitnorm_ood}
\begin{bchapter}
\bauthor{\bsnm{Wei}, \binits{H.}},
\bauthor{\bsnm{Xie}, \binits{R.}},
\bauthor{\bsnm{Cheng}, \binits{H.}},
\bauthor{\bsnm{Feng}, \binits{L.}},
\bauthor{\bsnm{An}, \binits{B.}},
\bauthor{\bsnm{Li}, \binits{Y.}}:
\bctitle{Mitigating neural network overconfidence with logit normalization}.
In: \bbtitle{{ICML}}.
\bsertitle{Proceedings of Machine Learning Research},
vol. \bseriesno{162},
pp. \bfpage{23631}--\blpage{23644}
(\byear{2022})
\end{bchapter}
\endbibitem

%%% 61
\bibitem[\protect\citeauthoryear{Wang et~al.}{2024}]{Wang2024tpami_cl}
\begin{barticle}
\bauthor{\bsnm{Wang}, \binits{L.}},
\bauthor{\bsnm{Zhang}, \binits{X.}},
\bauthor{\bsnm{Su}, \binits{H.}},
\bauthor{\bsnm{Zhu}, \binits{J.}}:
\batitle{A comprehensive survey of continual learning: Theory, method and application}.
\bjtitle{{IEEE} Trans. Pattern Anal. Mach. Intell.}
\bvolume{46}(\bissue{8}),
\bfpage{5362}--\blpage{5383}
(\byear{2024})
\end{barticle}
\endbibitem

%%% 62
\bibitem[\protect\citeauthoryear{Xu et~al.}{2024}]{xu2023scale_ood}
\begin{bchapter}
\bauthor{\bsnm{Xu}, \binits{K.}},
\bauthor{\bsnm{Chen}, \binits{R.}},
\bauthor{\bsnm{Franchi}, \binits{G.}},
\bauthor{\bsnm{Yao}, \binits{A.}}:
\bctitle{Scaling for training time and post-hoc out-of-distribution detection enhancement}.
In: \bbtitle{{ICLR}}
(\byear{2024})
\end{bchapter}
\endbibitem

%%% 63
\bibitem[\protect\citeauthoryear{Yu et~al.}{2020}]{yu2020semanticdrift_cl}
\begin{bchapter}
\bauthor{\bsnm{Yu}, \binits{L.}},
\bauthor{\bsnm{Twardowski}, \binits{B.}},
\bauthor{\bsnm{Liu}, \binits{X.}},
\bauthor{\bsnm{Herranz}, \binits{L.}},
\bauthor{\bsnm{Wang}, \binits{K.}},
\bauthor{\bsnm{Cheng}, \binits{Y.}},
\bauthor{\bsnm{Jui}, \binits{S.}},
\bauthor{\bsnm{Weijer}, \binits{J.}}:
\bctitle{Semantic drift compensation for class-incremental learning}.
In: \bbtitle{{CVPR}},
pp. \bfpage{6980}--\blpage{6989}
(\byear{2020})
\end{bchapter}
\endbibitem

%%% 64
\bibitem[\protect\citeauthoryear{Yan et~al.}{2021}]{yan2021dynamicER_cl}
\begin{bchapter}
\bauthor{\bsnm{Yan}, \binits{S.}},
\bauthor{\bsnm{Xie}, \binits{J.}},
\bauthor{\bsnm{He}, \binits{X.}}:
\bctitle{{DER:} dynamically expandable representation for class incremental learning}.
In: \bbtitle{{CVPR}},
pp. \bfpage{3014}--\blpage{3023}
(\byear{2021})
\end{bchapter}
\endbibitem

%%% 65
\bibitem[\protect\citeauthoryear{Yang et~al.}{2024}]{YangZ2024_ood}
\begin{barticle}
\bauthor{\bsnm{Yang}, \binits{J.}},
\bauthor{\bsnm{Zhou}, \binits{K.}},
\bauthor{\bsnm{Li}, \binits{Y.}},
\bauthor{\bsnm{Liu}, \binits{Z.}}:
\batitle{Generalized out-of-distribution detection: {A} survey}.
\bjtitle{Int. J. Comput. Vis.}
\bvolume{132}(\bissue{12}),
\bfpage{5635}--\blpage{5662}
(\byear{2024})
\end{barticle}
\endbibitem

%%% 66
\bibitem[\protect\citeauthoryear{Zhang et~al.}{2023}]{zhang2023she_ood}
\begin{bchapter}
\bauthor{\bsnm{Zhang}, \binits{J.}},
\bauthor{\bsnm{Fu}, \binits{Q.}},
\bauthor{\bsnm{Chen}, \binits{X.}},
\bauthor{\bsnm{Du}, \binits{L.}},
\bauthor{\bsnm{Li}, \binits{Z.}},
\bauthor{\bsnm{Wang}, \binits{G.}},
\bauthor{\bsnm{Liu}, \binits{X.}},
\bauthor{\bsnm{Han}, \binits{S.}},
\bauthor{\bsnm{Zhang}, \binits{D.}}:
\bctitle{Out-of-distribution detection based on in-distribution data patterns memorization with modern hopfield energy}.
In: \bbtitle{{ICLR}}
(\byear{2023})
\end{bchapter}
\endbibitem

%%% 67
\bibitem[\protect\citeauthoryear{Zenke et~al.}{2017}]{zenke2017si_cl}
\begin{bchapter}
\bauthor{\bsnm{Zenke}, \binits{F.}},
\bauthor{\bsnm{Poole}, \binits{B.}},
\bauthor{\bsnm{Ganguli}, \binits{S.}}:
\bctitle{Continual learning through synaptic intelligence}.
In: \bbtitle{{ICML}}.
\bsertitle{Proceedings of Machine Learning Research},
vol. \bseriesno{70},
pp. \bfpage{3987}--\blpage{3995}
(\byear{2017})
\end{bchapter}
\endbibitem

%%% 68
\bibitem[\protect\citeauthoryear{Zhou et~al.}{2024}]{Zhou202424_cil}
\begin{barticle}
\bauthor{\bsnm{Zhou}, \binits{D.}},
\bauthor{\bsnm{Wang}, \binits{Q.}},
\bauthor{\bsnm{Qi}, \binits{Z.}},
\bauthor{\bsnm{Ye}, \binits{H.}},
\bauthor{\bsnm{Zhan}, \binits{D.}},
\bauthor{\bsnm{Liu}, \binits{Z.}}:
\batitle{Class-incremental learning: {A} survey}.
\bjtitle{{IEEE} Trans. Pattern Anal. Mach. Intell.}
\bvolume{46}(\bissue{12}),
\bfpage{9851}--\blpage{9873}
(\byear{2024})
\end{barticle}
\endbibitem

%%% 69
\bibitem[\protect\citeauthoryear{Zhang et~al.}{2020}]{Zhang2020dmc_cl}
\begin{bchapter}
\bauthor{\bsnm{Zhang}, \binits{J.}},
\bauthor{\bsnm{Zhang}, \binits{J.}},
\bauthor{\bsnm{Ghosh}, \binits{S.}},
\bauthor{\bsnm{Li}, \binits{D.}},
\bauthor{\bsnm{Tasci}, \binits{S.}},
\bauthor{\bsnm{Heck}, \binits{L.P.}},
\bauthor{\bsnm{Zhang}, \binits{H.}},
\bauthor{\bsnm{Kuo}, \binits{C.-J.}}:
\bctitle{Class-incremental learning via deep model consolidation}.
In: \bbtitle{{WACV}},
pp. \bfpage{1120}--\blpage{1129}
(\byear{2020})
\end{bchapter}
\endbibitem

\end{thebibliography}

%% BioMed_Central_Bib_Style_v1.01

\end{document}